\documentclass[10pt,twocolumn,letterpaper]{article}

\usepackage{parskip}
\usepackage{times}
\usepackage{amsfonts,amsmath}
\usepackage{verbatim}
\usepackage[breaklinks=true,bookmarks=false]{hyperref}
\usepackage[font=small]{caption}

\usepackage{tabularray}
\usepackage{graphics}
\usepackage{graphicx}
\graphicspath{./figures/}

\usepackage{geometry}
\usepackage{adjustbox}

\usepackage[style=ieee]{biblatex}
\AtBeginBibliography{\small}

\newcommand{\eg}{\textit{e}.\textit{g}.\,}

\begin{document}

% ======================================================
% TITLE
% ======================================================
\title{\vspace{1cm}DeepATLAS: One-Shot Localization for Biomedical Data}

\author{\\
Peter D. Chang, MD\\
University of California Irvine, California, USA\\
{\tt\small changp6@hs.uci.edu}}

\date{}
\maketitle

% ======================================================
% ABSTRACT 
% ======================================================
\begin{abstract}

This paper introduces the DeepATLAS foundational model for localization tasks in the domain of high-dimensional biomedical data. Upon convergence of the proposed self-supervised objective, a pretrained model maps an input to an anatomically-consistent embedding from which any point or set of points (\eg boxes or segmentations) may be identified in a one-shot or few-shot approach. As a representative benchmark, a DeepATLAS model pretrained on a comprehensive cohort of 51,000+ unlabeled 3D computed tomography exams yields high one-shot segmentation performance on over 50 anatomic structures across four different external test sets, either matching or exceeding the performance of a standard supervised learning model. Further improvements in accuracy can be achieved by adding a small amount of labeled data using either a semisupervised or more conventional fine-tuning strategy. 

\end{abstract}

\section{Introduction}

Localization is a key task across high-dimensional biomedical datasets. In the domain of radiologic imaging, various efforts have focused on identifying, measuring, and/or quantifying anatomy across the body, such as the brain, lungs, heart, breast, liver, spleen, pancreas, kidney, prostate, and bones \cite{Altini2022LiverSurvey, Lenchik2019AutomatedReview, Fu2021ASegmentation}. In computer vision, localization is often formulated as an \textit{N}-class detection problem, whereby discrete objects of interest are identified using conventional semantic segmentation \cite{Ronneberger2015U-Net:Segmentation, Isensee2020NnU-Net:Segmentation}, bounding box \cite{Lin2020FocalDetection}, or landmark detection \cite{Noothout2020DeepImages} models. Such approaches however may be inefficient, requiring separate models for a potential infinite number of identifiable structures. As an alternative, this paper explores localization with a theoretical infinite number of classes (\(N \rightarrow \infty\)) correctly applied to every position in a dataset. Such a dense prediction map would represent a generalized solution to all localization tasks, as any single point or set of points can be used to identify regions of arbitrary granularity.

While the described task at first seems intractable, the inherent structural regularity of high-dimensional biomedical data allows for a viable solution. Indeed, compared to generic datasets with an open-ended number of detectable objects, medical device acquisitions often consist of views or poses of the same underlying structure: radiologic images reflect known anatomy of the human body, electrocardiograms interrogate standard waveforms of the cardiac cycle, esophageal manometry yields characteristic pressure changes of peristalsis during swallowing, and so on. Thus, though infinite, the labels of the dense mapping problem may be bounded by parameterizing all relevant structures with a coordinate space that spans the target domain. In this paper focused on radiologic computed tomography (CT) imaging, every position in the human body is mapped to a specific 3D coordinate in space from which the model may choose when labeling each discrete voxel with its underlying structure.  

As formulated, the proposed dense mapping problem is difficult to solve using standard supervised approaches as doing so requires manual annotation of every position with a predetermined, precise coordinate in space. Instead, an important contribution of this work is a self-supervised objective that fully defines a solution to this task without explicit annotations. This loss function is grounded on two key observations. First, consider that successful model convergence requires only \textit{consistent} labeling of any given anatomic structure without necessarily needing to know \textit{a priori} what the coordinate value itself is. Indeed, despite allowing the model to freely select coordinate values for any structure during optimization, label consistency may still be enforced by ensuring that image features at matching locations (identified by matching coordinate predictions) are highly correlated. Second, inherent anatomic consistency between exams implies that model-generated coordinate maps must exhibit reasonable smoothness and regularity. This criteria is enforced by minimizing the variance of the local Jacobian determinants across the coordinate map field. 

By learning to consistently label every position in an exam, the proposed foundational model efficiently generalizes to downstream localization tasks in a one-shot or few-shot manner. In prior work \cite{Shaban2017One-ShotSegmentation, Zhang2020SG-One:Segmentation}, conventional few-shot learning strategies for classification have been applied to semantic segmentation and other localization tasks. Recent advances have extended this to zero-shot segmentation in an interactive manner \cite{Mazurowski2023SegmentStudy, Huang2024SegmentImages}. However, these approaches are predicated upon pretraining with a standard \emph{supervised} objective to learn generalizable embeddings, a requirement that limits application to sparsely labeled biomedical datasets. Furthermore, self-supervised pretext tasks for natural images such as predicting rotation \cite{Gidaris2018UnsupervisedRotations}, solving jigsaws \cite{Noroozi2016UnsupervisedPuzzles}, or generative modeling \cite{Radford2015UnsupervisedNetworks} yield learned representations that only indirectly correlate to problems in the medical domain. Indeed, there is currently no unsupervised pretraining strategy with robust out-of-the-box one-shot performance on biomedical localization tasks.    

In this study, the proposed framework is used to generate foundational models using a large comprehensive cohort of radiology CT exams. After pretraining on 51,000+ unlabeled volumes across 100+ imaging protocols, a total of four external test sets are used to evaluate generalizability of the learned embeddings. An initial set of experiments evaluates one-shot segmentation and explores relative performance differences as a function of constrained pretraining cohort size. Subsequently, few-shot segmentation is assessed in the context of a joint loss function (simultaneous optimization of self-supervised and supervised objectives) as well as a standard fine-tuning procedure, both compared to a supervised baseline without pretraining. 

\begin{figure*}
    \centering
    \includegraphics[width=\textwidth]{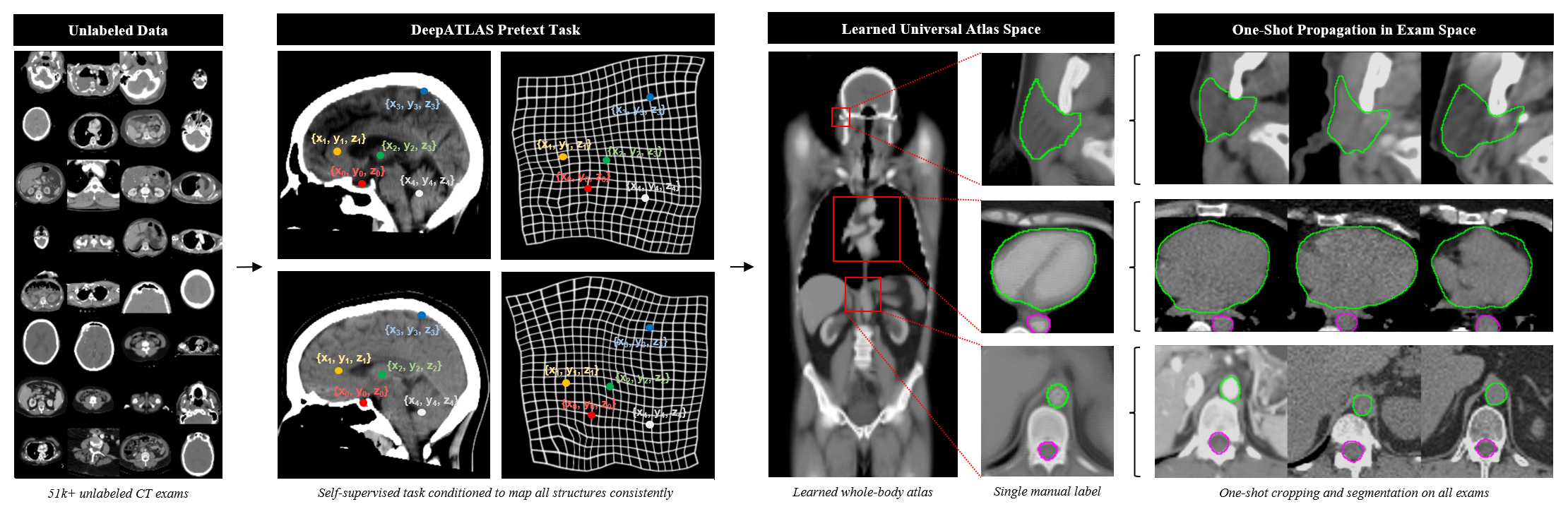}
    \caption{Overview of DeepATLAS Framework. A diverse, unlabeled cohort of 51k CT exams across multiple anatomic regions is used for unsupervised pretraining. The self-supervised pretext task is to label every point in an exam with a unique coordinate, requiring only that consistent coordinates are chosen for the same structure across all exams. Importantly, the contents of the coordinate space are not defined explicitly but rather learned by the model during optimization; after convergence, this derived coordinate space may be visually approximated by a model-reconstructed whole-body atlas. The anatomically-consistent embeddings generated by DeepATLAS may be used to propagate any reference localization task performed just once (such as cropping via bounding box or segmentation) on new exams without further training.} 
    \label{fig:overview}
\end{figure*}

\section{Methods}

\subsection{Overview}

This paper introduces DeepATLAS (Automorphic Task for Learning Anatomic Structure), a self-supervised framework that efficiently and generically solves localization tasks by learning anatomically-consistent embeddings---maps that densely encode the position of anatomic structures in a reliable manner across all exams (Figure \ref{fig:overview}). Importantly, the pretext task is conditioned on consistency alone such that any coordinate value may be chosen for a given structure so long as all instances are reliably encoded. Thus, the specific contents of the learned coordinate space are discovered independently during optimization as the model maps each encountered anatomic structure to a unique coordinate value. Upon convergence, the underlying structure encoded by any coordinate value may be visualized by a model-generated \textit{atlas} reconstruction, in turn reflecting the consistency, granularity, and quality of the derived feature embeddings (Figure \ref{fig:overview}).  

\begin{figure*}
    \centering
    \includegraphics[width=\textwidth]{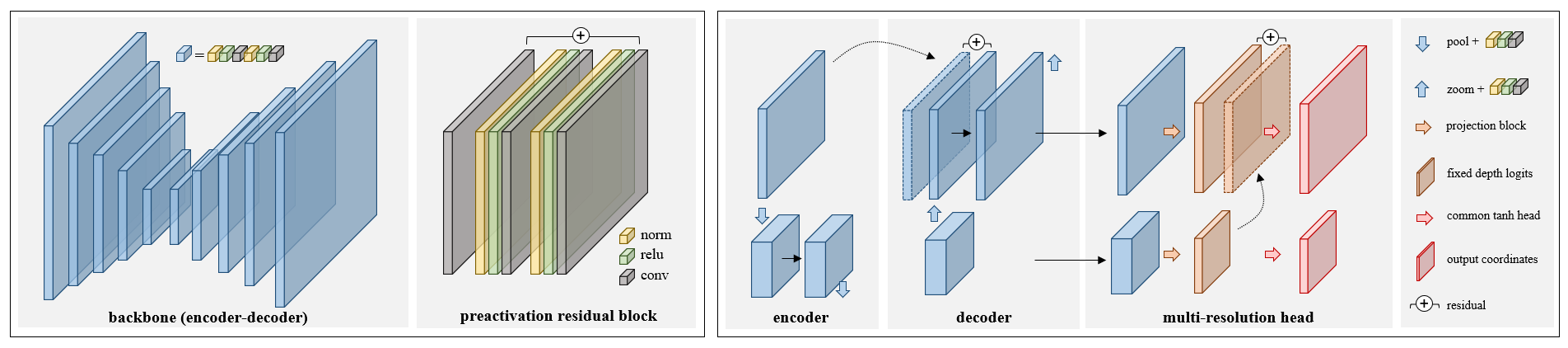}
    \caption{DeepATLAS Network Architecture.}
    \label{fig:model-00}
\end{figure*}

Let \(C \in \mathbb{R}^n\) represent a learned \(n\)-dimensional template coordinate space. For three-dimensional radiographic imaging data, \(C \in \mathbb{R}^3\) where each coordinate corresponds to an absolute location in the body: \((i, j, k)_{chiasm}\) may correspond to a single, unique coordinate for the optic chiasm, while \((i, j, k)_{carina}\) may correspond to a separate, unique coordinate for the carina. For any given exam \(x_n\), let \(c_n = F(x_n)\) represent an \emph{automorphic} mapping of the template coordinate space \(C \rightarrow c_n\) such that the underlying anatomy matches the patient-specific configuration present in \(x_n\): if present, the voxel containing the optic chiasm is labeled \((i, j, k)_{chiasm}\), while the voxel containing the carina is labeled \((i, j, k)_{carina}\). Here, the term \emph{automorphic} emphasizes that a single universal coordinate space is preserved between all transformations and that any single structure, if present, is labeled consistently (with the same coordinates) across all exams. In DeepATLAS, the automorphic function \(F(x_n)\) is implemented with a convolutional neural network (CNN), such that each forward-pass through the network performs a mapping \(C \rightarrow c_n\) satisfying the original input \(x_n\). 

\subsection{Network Architecture} \label{NetworkArchitecture}

An overview of the DeepATLAS network architecture is shown in Figure \ref{fig:model-00}. The network is implemented as a fully-convolutional encoder-decoder backbone \cite{Ronneberger2015U-Net:Segmentation} with preactivation ResNet-style blocks \cite{He2016IdentityNetworks} and multi-resolution dense prediction heads. Each preactivation convolutional block is defined as the serial application of activation normalization, a leaky ReLU (\(\alpha = 0.1\)) non-linearity, and a 3x3x3 convolutional kernel. Given small batch sizes during training, a combination of group normalization \cite{Wu2020GroupNormalization} and weight standardization \cite{Qiao2019Micro-BatchStandardization} is used instead of conventional batch normalization. Each \emph{residual block} is defined as a pair of preactivation \cite{He2016IdentityNetworks} convolutional blocks with a residual connection between successive convolutional layer outputs. 

The encoding arm of the backbone is initiated with a single convolutional operation applied to the input tensor followed by a single residual block. The remaining encoding arm is completed with alternating application of a single residual block followed by a parameterless downsampling operation repeated a total of \(L\) times, while the symmetric decoding arm reverses these operations with alternating application of a single residual block followed by a parameterless upsampling operation (see \ref{Resampling} for details on resampling strategy). In keeping with the ResNet-style design, skip connections between encoding and decoding layers are also implemented as residual operations.

\subsubsection{Multi-resolution Heads} At each resolution within the expanding arm, a dense regression head yields a total of \(c_l = \{c_0, c_1, c_2, ... c_L\}\) coordinate predictions, where \(c_0\) represents the native (full) resolution, \(c_1\) is the first subsample, and so on. While \(c_0\) represents the final (full-resolution)  prediction, the model is optimized sequentially starting with the most coarse \(c_L\) coordinate map first. Here, each incremental \(c_{l-1}\) coordinate map doubles the resolution and refines the prediction of the previous level. See \ref{Optimization} for further details regarding the optimization procedure.

An overview of the dense prediction head is shown if Figure \ref{fig:model-00}. At each level, the head is implemented as a single residual block (blue) followed by a single convolutional block (orange) yielding a fixed-depth (fixed-channel) feature map \(f_l\). Subsequently, a single shared convolutional block with a tanh activation function (same weights at all levels; red arrow) is applied to generate the final normalized coordinate predictions between \([-1, +1]\). At the deepest resolution level, the fixed depth feature map \(f_L\) is passed directly into the shared tanh-block head. At all other resolution levels, the current fixed depth feature map \(f_l\) is combined with the upsampled map from the previous level \(f_{l+1}\) using a residual operation before being passed into the shared tanh-block head. In this way, each multi-resolution head is designed to learn a residual feature map that refines predictions from the prior level. 

\subsection{Loss Formulation} \label{Loss Formulation}

The DeepATLAS pretext objective is optimized by evaluating coordinate prediction consistency through a pair of derived image registration tasks. Note that in this framework, these image registration tasks (and associated intermediate calculations) are used only to test the quality of coordinate predictions and to define a differentiable loss for optimization. After convergence, each model forward-pass is performed without any external dependencies or registration procedure. 

The first objective is termed the \emph{implicit} registration loss, a constraint that ensures locations between exams with the same coordinate prediction share similar image features. To calculate this objective, consider that overlapping regions between any pair of exams \(x_p\) and \(x_q\) may be mapped to one other with 1:1 correspondence by identifying matching values in the predicted coordinate spaces \(c_p\) and \(c_q\). More formally, a nonlinear warp field \(\Phi_{pq}\) that aligns volume \(x_p\) with \(x_q\) may be inferred through a sequential mapping of \(c_p \rightarrow C \rightarrow c_q\) as defined by:

\begin{equation} \label{eq:1}
    \Phi_{pq} = c_q \circ c_p^{-1}
\end{equation}

Here, the inverse warp \(c_p^{-1}\) maps the original volume \(x_p\) to the template coordinate space \(C\) while the subsequent forward warp \(c_q\) applies a second deformation to match the target volume \(x_q\). Note that this warp field for registration between two exams is not directly predicted by the CNN model, which is conditioned only to yield a single patient-specific coordinate space \(c_n\) for each exam \(x_n\). Nonetheless, any two exams may be aligned using equation \ref{eq:1} through a pair of corresponding \(c_p\) and \(c_q\) predictions. For the remainder the paper, the implicit warp field will be denoted as \(\Phi_I\). Interestingly, based on equation \ref{eq:1}, optimizing a registration task using \(\Phi_I\) has the favorable effect of encouraging smooth, invertible \(C \rightarrow c_n\) predictions. Indeed, as elucidated in ablation experiments (see \ref{Ablation Studies} for details), this feature inherent to the \(\Phi_I\) loss is a key contribution to high performance of the DeepATLAS framework. Finally, because the precise, true solution to nonlinear warp field inversion as outlined equation by \ref{eq:1} is numerically expensive, this paper proposes a novel, computationally efficient and fully differentiable approximation that yields a tractable solution for model training (see \ref{Implicit Warp} for details). 

The second objective is termed the \emph{explicit} registration loss. As previously noted, each model prediction yields a nonlinear warp field that maps \(C \rightarrow c_n\). Within the template coordinate space \(C\), let \(X\) represent a learnable atlas such that each \(c_n\) warp field also maps \(X \rightarrow x_n\). Functionally, the atlas \(X\) is similar to conventional population-based imaging atlases such as those defined in the Talairach \cite{Talairach1988Co-planarImaging} and MNI \cite{Collins1994AutomaticSpace.} coordinate spaces. However, rather than relying on a predefined template, the atlas \(X\) is optimized de novo as a single trainable tensor conditioned such that the deformation of \(X\) by \(c_n\) should yield an output that is similar to \(x_n\). Intuitively, this formulation allows the atlas \(X\) to adopt the most typical appearance of each position across the entire dataset. For the sake of consistency, the explicit warp field (which is equivalent to \(c_n\)) will be denoted \(\Phi_E\) for the remainder of the paper.

Through the \(\Phi_I\) and \(\Phi_E\) warp fields, the DeepATLAS self-supervised objectives can be formally defined as a pair of image registration tasks. Let \(\Phi x = x \circ \Phi\) represent the application of deformation field \(\Phi\) to \(x\). Additionally, let \(\mathcal{L}_{sim}(a, b)\) represent a function that estimates the similarity between \(a\) and \(b\). Finally, let \(\mathcal{L}_{smooth}(\Phi)\) represent a function that estimates the smoothness of a warp field \(\Phi\). Then, the implicit warp defines a registration task between a randomly sampled pair of exams (\(x_p \rightarrow x_q\)) such that:

\begin{equation} \label{eq:2}
    loss_{\Phi_I} \propto \mathcal{L}_{sim}(x_p \circ \Phi_I, x_q) + \mathcal{L}_{smooth}(\Phi_I)
\end{equation}

Additionally, the explicit warp defines a registration task between the learned template atlas and each exam (\(X \rightarrow x_n\)) such that:

\begin{equation} \label{eq:3}
    loss_{\Phi_E} \propto \mathcal{L}_{sim}(X \circ \Phi_E, x_n) + \mathcal{L}_{smooth}(\Phi_E)
\end{equation}

Together, these loss functions are optimized using popular strategies previously described for deep learning based registration and optical flow estimation \cite{Fu2020DeepReview, Balakrishnan2019VoxelMorph:Registration}. One key improvement is the use of feature-based similarity for implementation of \(\mathcal{L}_{sim}\) instead of more conventional distance-based metrics (mean absolute or squared error) or cross-correlation \cite{Kuppala2020AnApproaches}. As detailed in the appendix, \(\mathcal{L}_{smooth}\) estimates used to regularize warp field predictions are derived from the local Jacobian matrix determinant \cite{Kobyzev2021NormalizingMethods, DavidKerlick1982AssessingMatrix}.

\section{Results}

\begin{figure*}
    \centering
    \includegraphics[width=\textwidth]{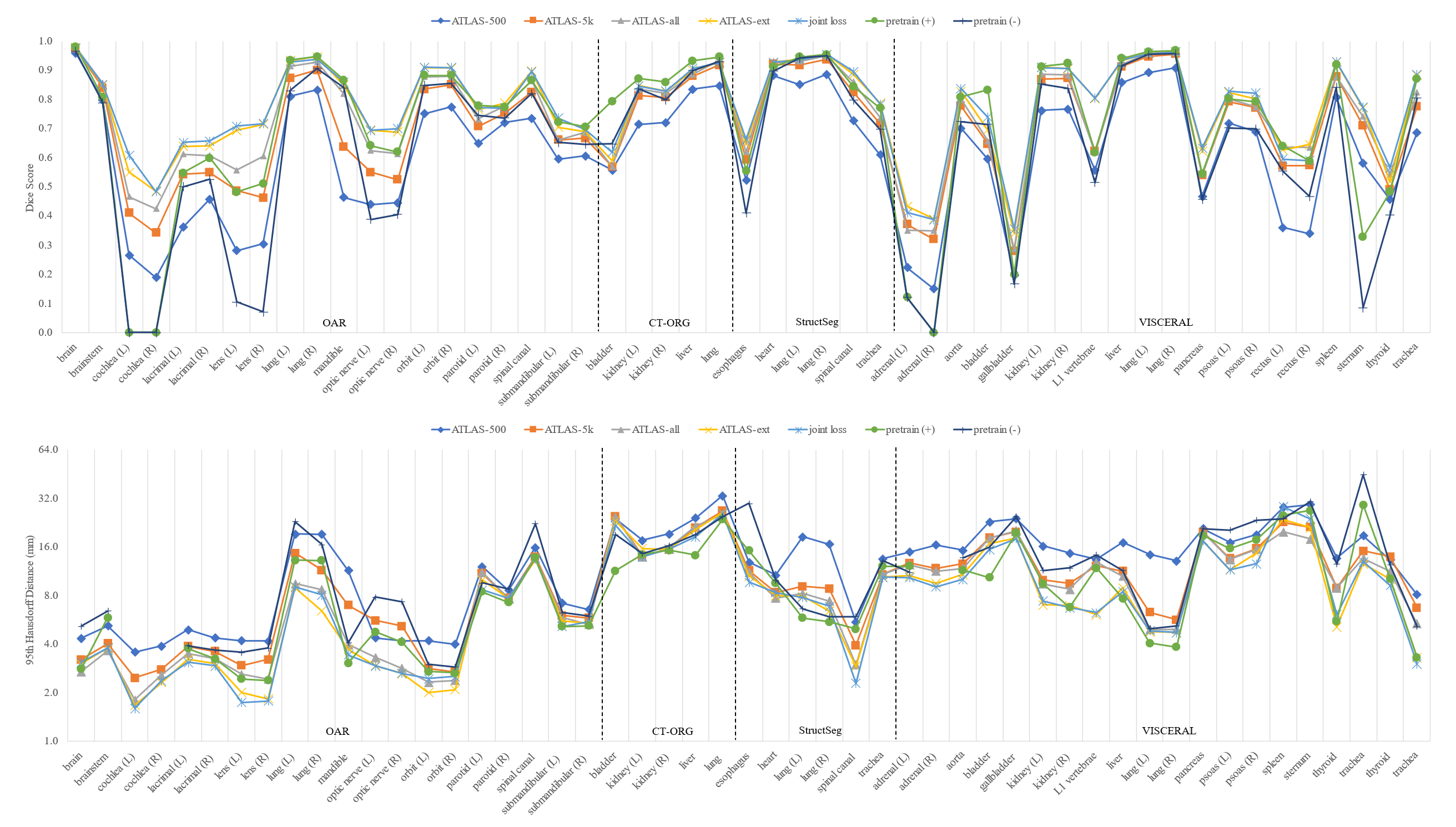}
    \caption{Summary of Dice Score and Hausdorff Distance Performance. Overall performance is plotted across all 51 anatomic structures and seven experiments for Dice score (top panel) and 95th percentile Hausdorff Distance (bottom panel). The baseline experiment (ATLAS-all) is compared to data-constrained (ATLAS-500 = 1\% of data, ATLAS-5k = 10\% of data), data-extended (ATLAS-ext = combined with external data), semisupervised (joint loss), as well as supervised (with and without pretraining) experiments. A total of four external test cohorts are included: (1) Head-and-Neck Organ-at-Risk (OAR); (2) CT Multi-Organ (CT-ORG); (3) Chest Organ-at-Risk (StructSeg); (4) Anatomy3 (VISCERAL).}
    \label{fig:stats-cmb}
\end{figure*}

\subsection{Datasets and Tasks}

The unlabeled dataset for self-supervised pretraining is comprised of 51,167 consecutive CT exams from a single academic medical center. The cohort is constructed from every diagnostic radiology CT exam obtained over a six month period from January 1, 2019 to June 30, 2019 spanning over 100 imaging protocols and heterogeneous anatomy including the head, face, orbits, sinus, temporal bones, neck, chest, heart, abdomen, pelvis, and spine (Table \ref{tab:data-distribution}). Furthermore, the cohort includes diverse acquisition parameters across matrix resolution, slice thickness, orientation, as well as all types and phases of contrast (intravenous, oral, and/or other injections). 

The labeled external dataset for evaluation is comprised of 265 CT exams from four different publicly-available, open-source datasets and includes segmentation masks for 51 anatomic structures. The Head-and-Neck Organ-at-Risk (OAR \cite{Nikolov2018DeepRadiotherapy}) dataset is comprised of 35 non-contrast CT exams and 20 labeled structures spanning the brain vertex to the mid-lungs. The CT Multi-Organ (CT-ORG \cite{Rister2020CT-ORGTomography}) dataset is comprised of 140 non-contrast and contrast-enhanced CT exams and 5 labeled structures spanning multiple field-of-views. The Chest Organ-at-Risk (StructSeg \cite{2019StructSegChallenge}) dataset is comprised of 50 contrast-enhanced CT exams and 6 labeled structures spanning the chest and upper abdomen. The Anatomy3 (VISCERAL \cite{Foncubierta2015Overview2015}) dataset is comprised of 40 non-contrast and contrast-enhanced CT exams and 20 labeled structures spanning neck, chest, abdomen, and pelvis. Together, the aggregate test set incorporates a diverse range of target anatomy including both large and small, subcentimeter structures (\eg lens and cochlea). Half of the labeled cohort is reserved, as needed, for experiments requiring annotated data including semisupervised and supervised model training. For ease of comparison, the remaining half of the labeled cohort is fixed as a test set for benchmarking across all experiments.

After model pretraining, segmentation masks are generated for the test set using one-shot, semisupervised, or supervised learning strategies. A combination of Dice score coefficient (DSC) and symmetric 95th-percentile Hausdorff distance (HD95) are used to assess model predictions relative to ground-truth. Comparisons between experiments are assessed using paired mean differences, with statistical significance calculated using a paired sample \textit{t}-test.

\subsection{One-Shot Label Propagation}

\subsubsection{Overview} In baseline experiments, the quality of DeepATLAS embeddings is assessed in the context of one-shot segmentation. After convergence of the self-supervised pretraining task, a single reference segmentation mask is generated for each target anatomic structure within the learned coordinate space. Subsequently, model-generated embeddings for each exam are used to propagate the reference segmentations using linear interpolation resampling. Note that in these experiments, no additional model fine-tuning is performed; all predictions are generated after self-supervised pretraining only. 

\begin{table*}[t]
    \caption{Pairwise Differences in Dice Score and Hausdorff Distance Across Experiments.}
    \label{table:ttest-diff}
    \begin{adjustbox}{width=\textwidth}
    \begin{tblr}{
      column{even} = {c},
      column{3} = {c},
      column{5} = {c},
      column{7} = {c},
      column{9} = {c},
      cell{1}{2} = {c=4}{},
      cell{1}{9} = {c=2}{},
      cell{3}{2} = {c=9}{},
      cell{11}{2} = {c=9}{},
      hline{1,19} = {-}{0.08em},
      hline{2} = {2-5,7,9-10}{},
      hline{3-4,11-12} = {-}{},
    }
    ~            & Unsupervised (One-Shot  Propagation) &                            &                            &                            & ~ & Semisupervised             & ~ & Supervised                 &                           \\
    ~            & ATLAS-500                              & ATLAS-5k                     & ATLAS-all                    & ATLAS-ext                    & ~ & joint loss                & ~ & pretrain (+)               & pretrain (-)              \\
    ~            & Dice Score                           &                            &                            &                            &   &                            &   &                            &                           \\
    ATLAS-500      & -                                    & \(+0.087\) \((p < 0.001)\) & \(+0.115\) \((p < 0.001)\) & \(+0.145\) \((p < 0.001)\) &   & \(+0.150\) \((p < 0.001)\) &   & \(+0.120\) \((p < 0.001)\) & \(+0.051\) \((p < 0.001)\) \\
    ATLAS-5k       & \(-0.087\) \((p < 0.001)\)           & -                          & \(+0.028\) \((p < 0.001)\) & \(+0.058\) \((p < 0.001)\) &   & \(+0.063\) \((p < 0.001)\) &   & \(+0.038\) \((p < 0.001)\) & \(-0.028\) \((p < 0.001)\) \\
    ATLAS-all      & \(-0.115\) \((p < 0.001)\)           & \(-0.028\) \((p < 0.001)\) & -                          & \(+0.030\) \((p < 0.001)\) &   & \(+0.035\) \((p < 0.001)\) &   & \(+0.010\) \((p = 0.005)\) & \(-0.053\) \((p < 0.001)\) \\
    ATLAS-ext      & \(-0.145\) \((p < 0.001)\)           & \(-0.058\) \((p < 0.001)\) & \(-0.030\) \((p < 0.001)\) & -                          &   & \(+0.005\) \((p < 0.001)\) &   & \(-0.018\) \((p < 0.001)\) & \(-0.078\) \((p < 0.001)\) \\
    joint loss  & \(-0.150\) \((p < 0.001)\)           & \(-0.063\) \((p < 0.001)\) & \(-0.035\) \((p < 0.001)\) & \(-0.005\) \((p < 0.001)\) &   & -                          &   & \(-0.023\) \((p < 0.001)\) & \(-0.083\) \((p < 0.001)\) \\
    pretrain (+) & \(-0.120\) \((p < 0.001)\)           & \(-0.038\) \((p < 0.001)\) & \(-0.010\) \((p = 0.005)\) & \(+0.018\) \((p < 0.001)\) &   & \(+0.023\) \((p < 0.001)\) &   & -                          & \(-0.070\) \((p < 0.001)\) \\
    pretrain (-) & \(-0.051\) \((p < 0.001)\)           & \(+0.028\) \((p < 0.001)\) & \(+0.053\) \((p < 0.001)\) & \(+0.078\) \((p < 0.001)\) & ~ & \(+0.083\) \((p < 0.001)\) & ~ & \(+0.070\) \((p < 0.001)\) & -                         \\
    ~            & Hausdorff Distance (mm)              &                            &                            &                            &   &                            &   &                            &                           \\
    ATLAS-500      & -                                    & \(-3.127\) \((p < 0.001)\) & \(-3.844\) \((p < 0.001)\) & \(-4.493\) \((p < 0.001)\) &   & \(-4.700\) \((p < 0.001)\) &   & \(-4.483\) \((p < 0.001)\) & \(-1.628\) \((p < 0.001)\) \\
    ATLAS-5k       & \(+3.127\) \((p < 0.001)\)           & -                          & \(-0.717\) \((p < 0.001)\) & \(-1.366\) \((p < 0.001)\) &   & \(-1.573\) \((p < 0.001)\) &   & \(-1.306\) \((p < 0.001)\) & \(+1.612\) \((p < 0.001)\) \\
    ATLAS-all      & \(+3.844\) \((p < 0.001)\)           & \(+0.717\) \((p < 0.001)\) &         -                  & \(-0.649\) \((p < 0.001)\) &   & \(-0.855\) \((p < 0.001)\) &   & \(-0.575\) \((p = 0.031)\) & \(+2.333\) \((p < 0.001)\) \\
    ATLAS-ext      & \(+4.493\) \((p < 0.001)\)           & \(+1.366\) \((p < 0.001)\) & \(+0.649\) \((p < 0.001)\) & -                          &   & \(-0.206\) \((p = 0.045\)) &   & \(+0.058\) \((p = 0.831)\) & \(+2.965\) \((p < 0.001)\) \\
    joint loss  & \(+4.700\) \((p < 0.001)\)           & \(+1.573\) \((p < 0.001)\) & \(+0.855\) \((p < 0.001)\) & \(+0.206\) \((p = 0.045)\) &   & -                          &   & \(+0.268\) \((p = 0.296)\) & \(+3.171\) \((p < 0.001)\) \\
    pretrain (+) & \(+4.483\) \((p < 0.001)\)           & \(+1.306\) \((p < 0.001)\) & \(+0.575\) \((p = 0.031)\) & \(-0.058\) \((p = 0.831)\) &   & \(-0.268\) \((p = 0.296\)) &   & -                          & \(+3.078\) \((p < 0.001)\) \\
    pretrain (-) & \(+1.628\) \((p < 0.001)\)           & \(-1.612\) \((p < 0.001)\) & \(-2.333\) \((p < 0.001)\) & \(-2.965\) \((p < 0.001)\) & ~ & \(-3.171\) \((p < 0.001)\) & ~ & \(-3.078\) \((p < 0.001)\) & -
    \end{tblr}
    \end{adjustbox}
\end{table*}

\subsubsection{Full Cohort} Figure \ref{fig:stats-cmb} and Tables \ref{table:summary-dsc}-\ref{table:summary-hd} summarize the results of one-shot label propagation after pretraining on the full 51k unlabeled dataset. Overall performance is robust across all 51 anatomic structures, with a mean aggregate DSC of 0.702 to 0.837 and HD95 of 4.9 to 20.2 mm across the various cohorts. Strong localization performance of small structures including 1-2 mm HD95 error for the cochlea and lens suggest high precision of the learned feature embeddings. Lowest performance is noted in organs with inherent poor visibility and/or subjective boundaries (\eg pancreas, adrenal glands), challenges that have been noted in previous studies using standard supervised approaches \cite{Yao2020AdvancesReview, Li2024MCNet:Images}.  

\begin{figure*}
    \centering
    \includegraphics[width=\textwidth]{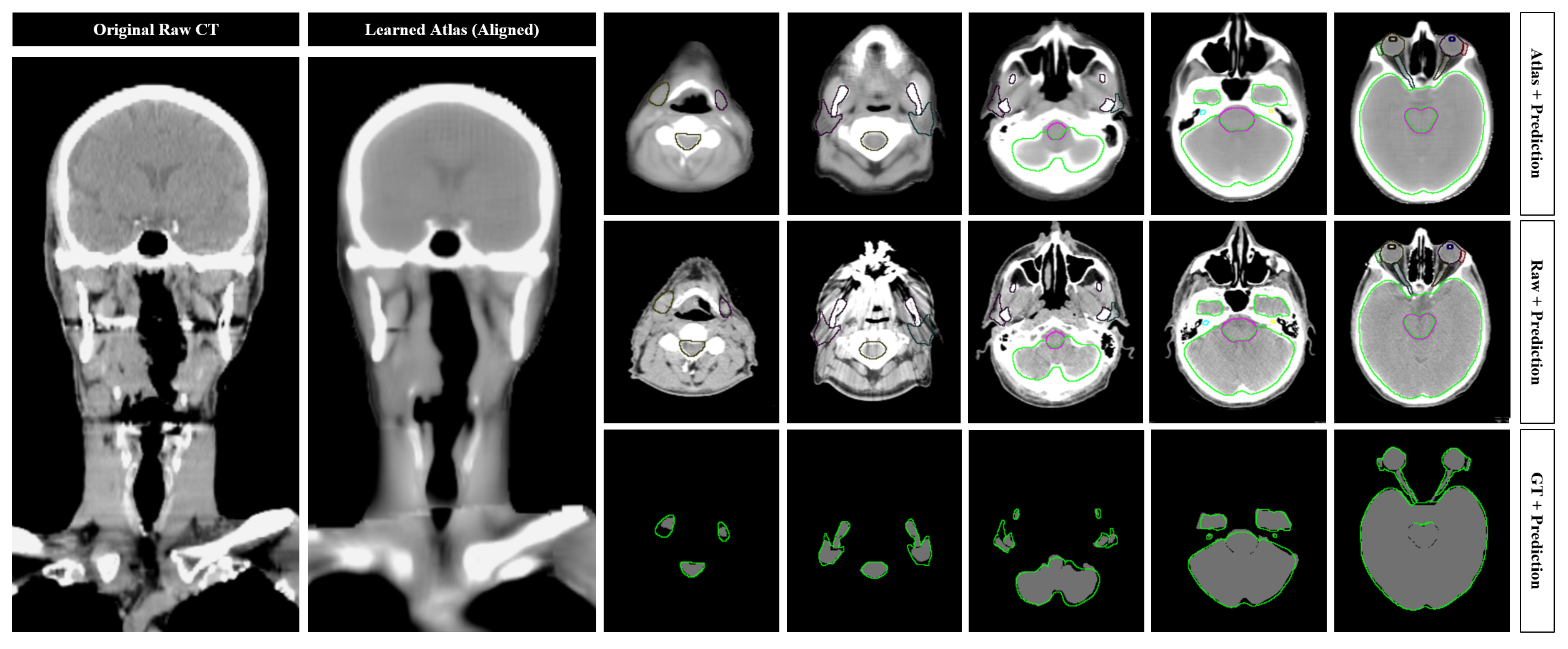}
    \caption{One-Shot Segmentation (Head-and-Neck Organ-at-Risk Cohort). A forward-pass of the DeepATLAS model labels every discrete position in an exam with a coordinate matching its underlying anatomy. To visually approximate the predicted anatomy at each location, the generated coordinate map may be used to project the learned atlas (a reconstruction of the learned coordinate space) to any given exam. As shown in the left two panels, the aligned atlas reconstruction exhibits high-fidelity correlation with raw CT data, suggesting the ability to generalize across many anatomic structures including those not directly evaluated in this experiment. In the right panels, single-shot segmentation masks are shown overlaid on the aligned atlas (top panels), raw CT data (middle panels), and ground-truth (bottom panels). In this example, segmentation masks are shown for the lens, orbit, optic nerve, lacrimal gland, cochlea, mandible, parotid gland, submandibular gland, brain, and brainstem.}
    \label{fig:oar}
\end{figure*}

\begin{figure*}
    \centering
    \includegraphics[width=\textwidth]{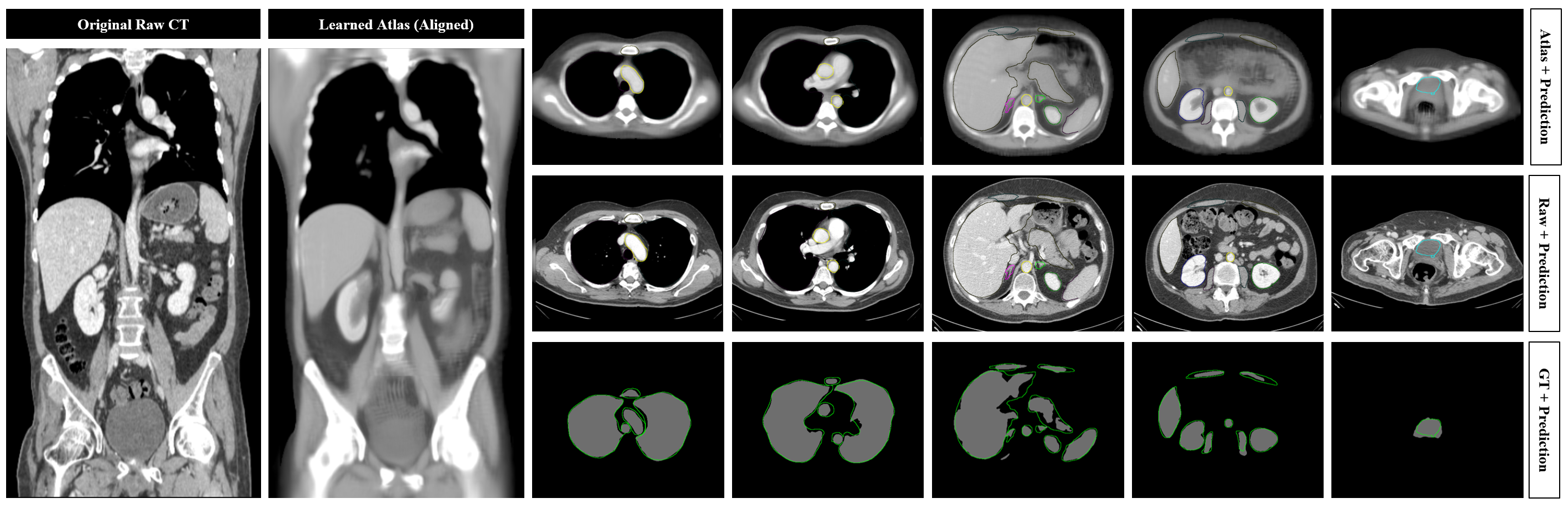}
    \caption{One-Shot Segmentation (Anatomy3 Cohort). In this example, segmentation masks are shown for the lungs, trachea, sternum, thoracic aorta, abdominal aorta, liver, spleen, pancreas, kidneys, iliopsoas muscle, and bladder. See Figure \ref{fig:oar} for further details.}
    \label{fig:visceral}
\end{figure*}

As a method to visualize and evaluate the overall consistency of model estimates across the entire volume, the predicted coordinates are used to project the learned atlas into alignment with each exam (Figures \ref{fig:oar} and \ref{fig:visceral}). Visualization of these aligned atlas reconstructions show high-fidelity correlation with raw CT data, suggesting the ability to generalize across many other additional anatomic structures beyond those directly evaluated in this experiment. 

\subsubsection{Constrained Cohorts} To assess incremental gain in model performance as a function of dataset size, the unlabeled dataset for algorithm pretraining is constrained on a logarithmic scale to include 1\% (approximately 500 exams; ATLAS-500) and 10\% (approximately 5,000 exams; ATLAS-5k) of the full cohort. Aside from this modification, all other implementation details are identical to the initial one-shot baseline experiment.

Figure \ref{fig:stats-cmb} and Tables \ref{table:summary-dsc}-\ref{table:summary-hd} summarize the results of one-shot label propagation after pretraining on various constrained unlabeled datasets. Table \ref{table:ttest-diff} summarizes the incremental mean differences in performance with each experiment. For each entry, the mean difference is calculated by subtracting the experiment designated by row from the experiment designated by column. For example, the difference between ATLAS-5k and ATLAS-500 is calculated at \(+0.087\) DSC and \(-3.127\) mm HD95. Overall, statistically significant improvements in DSC and HD95 are noted with increased pretraining dataset size. The magnitude of gain is most noticeable from ATLAS-500 to ATLAS-5k (\(+0.087\) DSC and \(-3.127\) mm HD95; \(p < 0.001\)) while the effect is more modest from ATLAS-5k to ATLAS-all (\(+0.028\) DSC and \(-0.717\) mm HD95, \(p < 0.001\)).  

\subsection{Semisupervised Learning}

\subsubsection{Combined Cohorts} Poor generalizability is a commonly-recognized limitation of many deep learning models in the medical domain \cite{Zhou2021APromises}. While this in part relates to constraints in available dataset size, another significant barrier is the relative lack of heterogeneity when training from single-site cohorts. A model trained primarily through a self-supervised objective offers a potential unique solution---without the need for manual annotations, data from arbitrary external sources can be easily included in pretraining to ensure maximum generalizability for target populations.

To assess this potential effect, raw CT imaging from the external cohorts (without labels) is added to the original full internal cohort to yield a combined dataset (ATLAS-ext) for self-supervised pretraining. As before, all other implementation details are identical to the previous experiments for one-shot label propagation.

Figure \ref{fig:stats-cmb} and Tables \ref{table:summary-dsc}-\ref{table:summary-hd} summarize the results of one-shot label propagation after pretraining on the combined internal-external dataset. Table \ref{table:ttest-diff} summarizes the incremental mean differences in performance from baseline experiments. Overall, statistically significant improvements in both DSC and HD95 are noted over the baseline experiment pretrained on the internal cohort alone (\(+0.030\) DSC and \(-0.649\) mm HD95; \(p < 0.001\)). It is interesting to note that the magnitude of improvement is similar to that observed in the previous experiments when scaling from 5k to 51k+ exams, suggesting that the increased diversity afforded by just several hundred external exams is comparable to the incremental benefit of 10x increased data from a single homogeneous source.

In addition to one-shot baseline performance, a series of experiments are designed to evaluate the incremental benefit of semisupervised learning using a portion of the available labeled data. In each of the following experiments, approximately half of the external cohort (between 17-70 labeled exams across each of the four datasets) is used for training, while the remaining half is used for testing (identical to the cohort used in previous one-shot experiments). Note that the relatively small number of labeled training data in these experiments is much lower than typical state-of-the-art supervised segmentation models commonly optimized on hundreds or thousands of annotated cases.

\subsubsection{Joint Loss Function} As an extension of the fully self-supervised pretext task,  annotations (when available) may be incorporated as an additional auxiliary label component to the loss function. At baseline, the pretext task is formulated such that anatomic consistency is defined by features uniformly distributed throughout image space. To add supervisory signal from annotated data, consistency can be further conditioned in label space (\eg all points within one mask must correspond to all points in another mask); practically, this is implemented simply by concatenating additional channels to the learned atlas (see \ref{Auxillary Label Loss} for more details). By simultaneously optimizing both the self-supervised and auxiliary supervised objectives, the joint loss limits any potential overfitting or other degradation in performance that may arise from standard supervised training on small datasets.

Figure \ref{fig:stats-cmb} and Table \ref{table:ttest-diff} summarize the results of semisupervised learning implemented using the described joint optimization method. Overall, modest but statistically significant improvements in both DSC and HD95 are noted over both the ATLAS-all (internal-only; \(+0.035\) DSC and \(-0.855\) mm HD95; \(p < 0.001\)) and ATLAS-ext (internal-external combined; \(+0.005\) DSC and \(-0.206\) mm HD95; \(p = 0.045\)) experiments. Of note, while overall DSC improves across the four external cohorts, the effect on HD95 is dampened and in fact slightly worsens in two of the cohorts. As HD95 is a metric that is more sensitive to outliers than DSC, this suggests that improved spatial overlap is achieved at a trade-off for small regions of increased divergence along the mask contours.

\subsubsection{Supervised Fine-Tuning} While joint optimization of the self-supervised and supervised objectives helps to stabilize model convergence and generalizability, the regularizing effect of optimizing both loss components may limit the theoretical learning potential of a fully supervised approach. Furthermore, simultaneous optimization requires access to the full unlabeled cohort and significant computational resources, both of which may preclude ease of reuse. Acknowledging these barriers, pretrained models are assessed for utility in a more conventional supervised fine-tuning paradigm. In this experiment, the baseline model pretrained on ATLAS-all is used to initialize weights for a supervised model after replacing the last convolutional layer with a new two-layer segmentation head. The entire model is then optimized across all layers using available labeled data only.

Figure \ref{fig:stats-cmb} and Table \ref{table:ttest-diff} summarize the results of supervised fine-tuning using weights initialized by ATLAS-all. Overall, modest but statistically significant improvements in both DSC and HD95 are noted over the baseline one-shot model (\(+0.010\) DSC, \(p = 0.005\); \(-0.575\) mm HD95, \(p = 0.031\)). In aggregate, supervised fine-tuning is inferior to the previous joint optimization strategy (-0.023 DSC and +0.268 mm HD95, \(p < 0.001\)), however this effect is likely primarily driven by the small number of available labeled exams for training. As evidence of this, supervised fine-tuning on the CT-ORG dataset (containing N=70 exams for training, the largest of any external dataset) was demonstrably superior to the joint optimization technique (\(+0.056\) DSC and \(-3.318\) mm HD95, \(p < 0.001\)), and in fact yielded the best results across any experiment. Overall, these findings suggest that the joint optimization technique may be a superior choice for few-shot learning while the supervised fine-tuning approach is preferred as the availability of labeled data increases.

\subsubsection{Supervised Baseline} As a baseline comparison, a standard supervised model is optimized without self-supervised pretraining. This model is identical in architecture to the previous network used for supervised fine-tuning (including a standard segmentation head) and optimized de novo from random weights using only the available labeled training cohort (no unlabeled data). 

Figure \ref{fig:stats-cmb} and Table \ref{table:ttest-diff} summarize the results of optimization with the supervised loss alone. Compared to random initialization, pretrained weights significantly improve model performance in all external cohorts by a large margin (\(+0.070\) DSC and \(-3.078\) mm HD95). In comparison to all other previous experiments, the supervised model outperforms only the constrained ATLAS-500 one-shot model but is significantly inferior to all others. Of note, the relatively small number of labeled exams used for training markedly limits the accuracy of a supervised learning model, and in fact fails to converge for several very small structures (\eg no predictions are generated for the right or left cochlea in any patient).

\section{Discussion}

\begin{figure*}
    \centering
    \includegraphics[width=\textwidth]{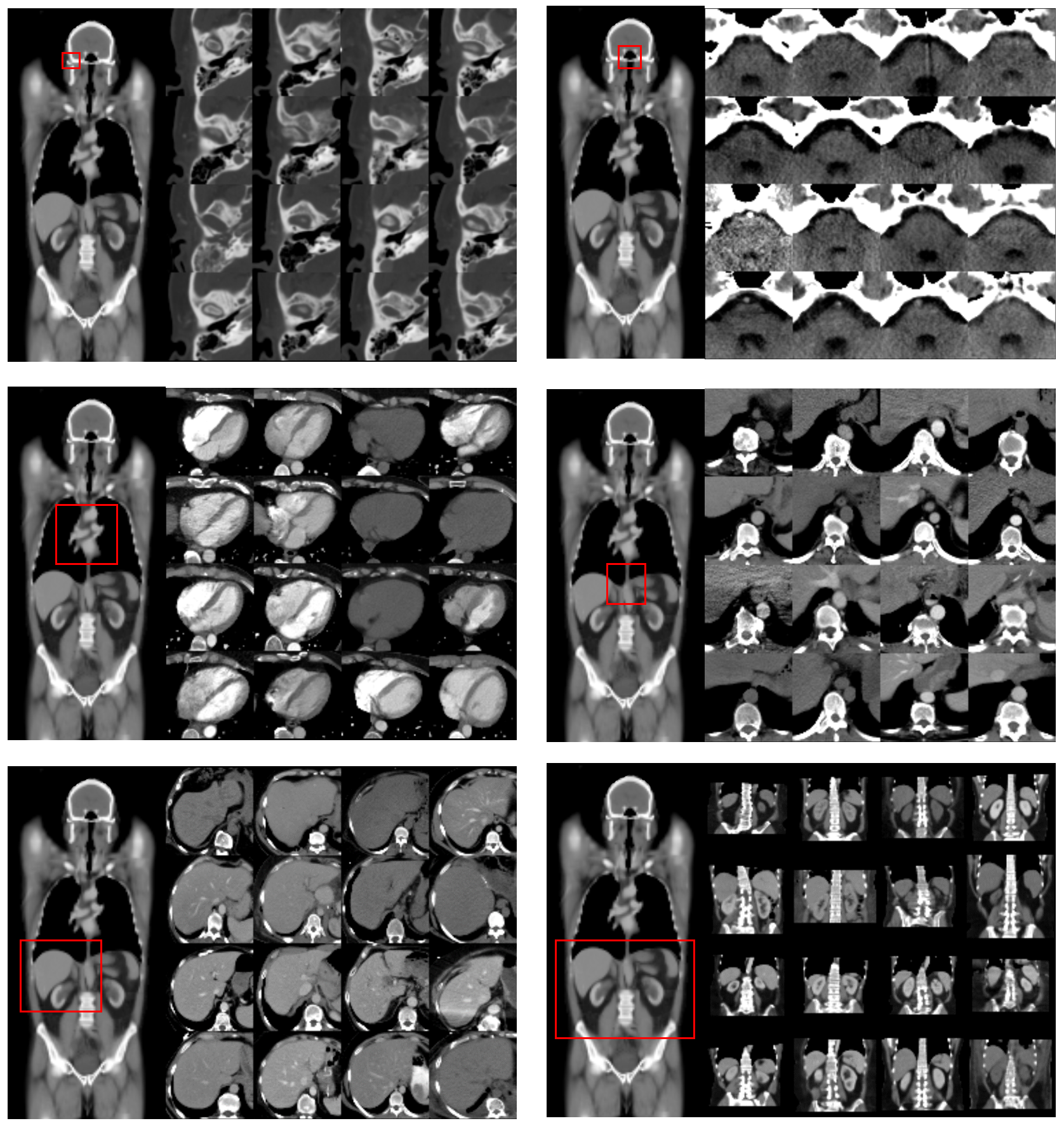}
    \caption{DeepATLAS for Automated Data Preprocessing. Any template bounding cube defined once in the learned atlas space can be used to query for exams containing the target region-of-interest. Furthermore, matching exams can be cropped and resampled to a uniform standard field-of-view and orientation (matching the coordinates of the template bounding cube). In these examples, representative fully automated results are shown for the right temporomandibular joint, pons, heart, diaphragmatic hiatus, right upper quadrant, and mid-abdomen.}
    \label{fig:cropping}
\end{figure*}

The self-supervised DeepATLAS framework yields a flexible, generalized representation of anatomic structure with the capacity for one-shot or few-shot localization on high-dimensional biomedical data. While supervised models for localizing anatomy have become increasingly prevalent and comprehensive in recent years \cite{Fu2021ASegmentation, Wasserthal2023TotalSegmentator:Images, Ma2021AbdomenCT-1K:Problem}, there is inevitably a need to solve for new tasks not already addressed by existing algorithms. In this context, the proposed DeepATLAS strategy is a much more efficient and scalable alternative, with reasonable one-shot generalizability for most problems. In addition, conventional segmentation or bounding box models provide only coarse estimates of structure compared to granular DeepATLAS predictions, which in turn support more precise anatomic characterization such as orientation or internal structure. For example, instead of simply identifying the bounds of an anatomic target, DeepATLAS may further reorient the region of interest into a standardized, consistent projection (Figure \ref{fig:cropping}). Moreover, compared to recent zero-shot interactive segmentation algorithms \cite{Mazurowski2023SegmentStudy, Huang2024SegmentImages}, DeepATLAS does not require identifying seed regions or initial predictions on an exam-by-exam basis. Finally, as a domain-agnostic unsupervised paradigm, the proposed methods may be easily extended to various other biomedical datasets. 

Compared to standard supervised models targeting a single or few target structures, DeepATLAS simultaneously labels every discrete position in an exam. The inherent global context of each prediction is particularly advantageous when morphology and other local cues alone are insufficient for robust performance, for example, when relative position (\eg the adrenal glands are primarily identified based on location above the kidneys) or long-range dependencies (\eg the L1 vertebrae is primarily identified as the first non-rib bearing vertebrae) are critical. Furthermore, the interdependence of each coordinate estimate suggests that predictions are strongly conditioned on an expected distribution of anatomy as learned during pretraining. As evidence of this, DeepATLAS predictions appear to be relatively robust to noise and other inherent sources of imaging heterogeneity but may be limited when there is significant deviation from the expected anatomy (\eg missing and/or variant anatomy, or randomly distributed pathological entities). For such tasks, a supervised loss component and/or fine-tuning is required for robust performance. Of note, this regularizing effect of the learned prior predisposes to more consistent, though at times imperfect, predictions. Indeed, while the addition of supervised loss components may improve overall spatial overlap based on DSC metrics, fully unsupervised models sometimes yielded stronger performance on outlier-sensitive HD95 metrics. 

A key finding from these experiments is the relative significant gain in model performance by simply adding target external cohorts to the pretraining pipeline, an improvement similar to that observed after a magnitude order increase in training cohort size. This suggests the potential for improved site-specific generalizability without the need for manual annotations on new data \cite{Navarro2022Self-SupervisedPerspective}. Extending this formulation, the DeepATLAS loss itself may be interpreted as a proxy for data consistency: any exam with a high loss value is one that diverges from the learned representation and may be prioritized for retraining. Such an approach may be further leveraged in the context of out-of-distribution detection \cite{Bulusu2020AnomalousSurvey} and active learning \cite{Budd2021AAnalysis}. 

Given the semantic importance of structure in the interpretation of biomedical data, the DeepATLAS learned representation may be used for a number of practical tasks. In addition to segmentation and other localization objectives, perhaps the most salient application is dataset preprocessing, a critical bottleneck in many development pipelines \cite{Willemink2020PreparingLearning}. As noted in \ref{Cascaded}, consider that DeepATLAS embeddings may be used to interrogate the presence (or percentage coverage) of any defined target anatomy and to prepare data volumes spanning the region of interest in a standard view (Figure \ref{fig:cropping}). Additional applications to dataset curation include: metadata extraction (\eg detecting phase of contrast administration by measuring density in target structures); semi-automated annotation; linear or deformable coregistration. Additionally, DeepATLAS feature embeddings may be used to improve downstream modeling tasks through the explicit incorporation of anatomic context. Several examples include: de novo training (\eg concatenating feature embeddings to raw data as model input); content-based image retrieval \cite{Saritha2019ContentProcess} (\eg identifying similar features while accounting for specific anatomy); anomaly detection \cite{Fernando2021DeepSurvey} (\eg identifying outlier features while accounting for specific anatomy); change detection \cite{Mandal2022AnNeeds} (\eg identifying structural changes or shifts between exams). 

\printbibliography

\clearpage

\section{Appendix}

\subsection{Loss Formulation}

The DeepATLAS pretext objective is optimized by evaluating coordinate prediction consistency through a pair of derived image registration tasks (implicit \(\Phi_I\) and explicit \(\Phi_E\) warps). Together, these tasks ensure that: (1) locations between exams with the same coordinate prediction share highly correlated image features (similarity loss); and (2) model-generated coordinate maps exhibit topology-preserving regularity (smoothness loss). 

\subsubsection{Implicit Warp} \label{Implicit Warp}

The first registration task used for optimization is an exam-to-exam alignment defined by the implicit warp \(\Phi_I\). As described above, \(\Phi_I\) is defined for each pair of model predictions \(c_p\) and \(c_q\) based on equation \ref{eq:1}. In this manner, each pair of exams \(x_p\) and \(x_q\) is related to one other by:

\begin{equation} \label{eq:4}
    x_p \circ \Phi_I \sim x_q
\end{equation}

While precise solution to equation \ref{eq:1} is numerically expensive, a recursive approximation to \(\Phi_I\) can be defined as a series of warp compositions. Let \(D_q\) represent the direction cosine matrix of coordinate system \(c_q\); for a nonlinear grid of coordinates, \(D_q\) is estimated by taking the mean of all local Jacobian matrices across all positions in \(c_q\). Additionally, let \(c_p\) represent the moving coordinate system where \(c_p \circ \Phi_I = c_q\). Then, to a first approximation \(t_0\), the warp field aligning \(c_p \rightarrow c_q\) can be estimated by:

\begin{equation} \label{eq:5}
   \Phi_I(t_0) = (c_q - c_p) \cdot D_q 
\end{equation}

Application of this initial warp field estimate \(\Phi_I(t_0)\) to \(c_p\) yields \(c_p' = c_p \circ \Phi_I(t_0)\), an intermediate, partially-transformed coordinate system (\(c_p \rightarrow c_p' \rightarrow c_q\)). At this point, the intermediate \(c_p'\) may replace the original \(c_p\) in equation \ref{eq:5} to yield \(\Phi_I(\delta_0) = (c_q - c_p') \cdot D_q\) where all other terms are identical. Here, \(\Phi_I(\delta_0)\) represents an approximation to the incremental difference that completes the mapping \(c_p' \rightarrow c_q\), and thus the updated estimate of \(\Phi_I(t_1) = \Phi_I(t_0) + \Phi_I(\delta_0)\). More generally, the recursive function may be defined with the following two equations:

\begin{equation} \label{eq:6}
    \Phi_I(t_n) = \Phi_I(t_{n-1}) + \Phi_I(\delta_{n-1})
\end{equation}

\begin{equation} \label{eq:7}
    \Phi_I(\delta_{n-1}) = (c_q - c_p \circ \Phi_I(t_{n-1})) \cdot D_q
\end{equation}

To initiate the iterative calculation, equation \ref{eq:7} is seeded with \(\Phi_I(t_0)\) as defined by equation \ref{eq:5}. Convergence is predicated on the assumption that the coordinate system \(c_q\) is valid (has a nonzero contribution to the standard basis in Cartesian notation and thus a well-defined \(D_q\)) and locally linear (differentiable). If these conditions are met, then each iterative update \(\Phi_I(\delta_{n-1})\) represents an incrementally smaller step. Upon convergence, \(c_p \circ \Phi_I(t_\infty) = c_q\) and thus the update defined in equation \ref{eq:7} is zero. 

Through empiric testing, a total of \(t = 20\) recursive updates is sufficient for subvoxel precision (\eg error in equation \ref{eq:7} to be less than 1.0) at over 0.99+ of all voxel locations across multiple experiments. It is further noted that this method is highly efficient and fully differentiable, requiring only a series of linear interpolation resampling operations for implementation.

\subsubsection{Explicit Warp}

The second registration task used for optimization is an atlas-to-exam alignment defined by the explicit warp \(\Phi_E\). As described above, \(\Phi_E\) is equivalent to the \(C \rightarrow c_n\) mapping performed by each forward-pass for the model. In this manner, the learnable atlas \(X\) in the universal template space \(C\) is related to each specific exam \(x_n\) by:

\begin{equation} \label{eq:8}
    X \circ \Phi_E \sim x_n 
\end{equation}

The atlas \(X\) is implemented simply as a single learnable tensor independent of the primary model backbone. As both \(X\) and \(\Phi_E\) are fully differentiable, the optimization process simultaneously learns: (1) an atlas \(X\) that best approximates each \(x_n\) after warping, as well as (2) a model that generates the best warp \(\Phi_E\) to align the atlas \(X\) to each \(x_n\). To improve the speed and stability of convergence, the atlas \(X\) may be seeded with a random exam that spans the target template space. Note that the atlas \(X\) can be trained at any resolution as the resampling operator can also perform any necessary downsampling or upsampling to match the target shape of \(x_n\). In these experiments, the atlas \(X\) is fixed at double the resolution of the original data.

\subsubsection{Similarity Loss}

In both coregistration tasks, similarity between the transformed (moving) \(x_m\) and the target (fixed) \(x_f\) volumes is determined using a feature-based metric. Specifically, the loss is calculated as the normalized cross-correlation between \(x_m\) and \(x_f\) within feature space derived from a pretrained autoencoder. 

Let \(R = \{r_0, r_1, r_2, ..., r_L\}\) define a total of \(L\) intermediate representations at different resolutions generated through a pretrained autoencoder, where each representation in \(R\) is a dense feature map spanning \(c\) channels. Let \(R_m\) and \(R_f\) represent the corresponding feature maps generated for each \(x_m\) and \(x_f\) volume, respectively. Then, the normalized cross-correlation (product moment correlation coefficient) is defined as:

\begin{equation}
    \sum_{l=0}^{L}(\frac{\sum(R_m - \bar{R}_m)(R_f - \bar{R}_f)}{\sqrt{\sum(R_m - \bar{R}_m)^2\sum(R_f - \bar{R}_f)^2}})
\end{equation}

In general, this formulation of similarity over a potential range of \(L\) feature resolutions defines a flexible loss that may incorporate a combination of low-level statistics and high-level abstract features as needed. By comparison, a simple distance-based error loss captures only single voxel-level information; in ablation experiments, a mean squared error (MSE) similarity loss is significantly inferior to the feature-based formulation (see \ref{Ablation Studies} for details). In these experiments, a total of two feature resolutions \(L = \{0, 1\}\) are used in all similarity losses. See \ref{Autoencoder} for implementation of autoencoder network architecture.

\subsubsection{Smoothness Loss}

In both coregistration tasks, the warp fields \(\Phi_I\) and \(\Phi_E\) are further conditioned with a smoothness constraint to encourage topology-preserving, continuous mappings. The smoothness loss is comprised of two components, both derived from the local Jacobian matrix determinant across each warp field.

Let \(J_p\) represent the local Jacobian matrix at point \(p\), and let \(\hat{J}_p = [\hat{j_0} \ \hat{j_1} \  \ldots \ \hat{j_m}]\) represent the corresponding column normalized local Jacobian matrix. Furthermore, let \(D_p = \log(det(J_p))\) and \(\hat{D}_p = \log(det(\hat{J}_p))\) represent the log-transform of the determinant of each corresponding Jacobian matrix, and \(\bar{D}_p\) represent the global mean of \(D_p\) across the coordinate field. For any given warp field \(\Phi\), \(D_p \rightarrow \bar{D}_p\) promotes warp field uniformity and thus a common strategy to enforce warp smoothness is to minimize the variance of \(D_p\) across all points \cite{Kobyzev2021NormalizingMethods}. Related to this strategy, \(\hat{D}_p \rightarrow 1\) promotes orthonormal columns in the Jacobian matrix and is used as a component of the smoothness loss to encourage locally-rigid transformations.

Both definitions above may be implemented by minimizing the L1 or L2 distance to the target reference, however to stabilize the optimization process, a Huber loss formulation \cite{Huber1964RobustParameter} is used instead:

\begin{equation}
    L_{\delta}(a) = 
    \begin{cases}
        \frac{1}{2}a^2                      & \text{if } |a| < \delta, \\
        \delta(|a| - \frac{1}{2}\delta)     & \text{otherwise.}
    \end{cases}
\end{equation}

By exhibiting quadratic behavior for small values of \(a\) and linear behavior for larger values, the Huber loss formulation is less sensitive to outliers. For all experiments, the Huber loss \(\delta = 0.5\). Thus, the full formulation of the smoothness loss is:

\begin{equation}
    \mathcal{L}_{smooth} = L_{0.5}(D_p - \bar{D}_p) + L_{0.5}(\hat{D}_p - 1)
\end{equation}

\subsubsection{Auxiliary Label Loss} \label{Auxillary Label Loss}

By default, \(\mathcal{L}_{sim}\) defines similarity based on cross-correlation of autoencoder-derived feature maps \(R\). As needed, additional formulations of similarity, defined through manually-engineered features and/or labeled data \cite{Fu2020DeepReview, Balakrishnan2019VoxelMorph:Registration}, may be used to further constrain the \(\mathcal{L}_{sim}\) component of both the implicit and explicit warp losses. Let \(x'\) and \(X'\) represent the concatenation of the original exam \(x\) and the learned atlas \(X\) with any additional auxiliary conditional features as needed. Then, equation \ref{eq:4} may be rewritten as:

\begin{equation}
   x'_p \circ \Phi_I \sim x'_q 
\end{equation}

Similarly, equation \ref{eq:5} may be rewritten as:

\begin{equation}
    X' \circ \Phi_E \sim x'_n\
\end{equation}

Here, similarity along the first channel of each concatenated representation remains defined by cross-correlation of autoencoder-derived features. The auxiliary information concatenated along the remaining channels however are encoded as low-level representations and thus matched using a standard MSE loss. In these experiments, the image gradient for all exams as defined by a simple Prewitt operator \cite{Prewitt1970ObjectExtraction} is used as an additional auxiliary map to maximize correspondence of edge features. In addition, for experiments using joint optimization of self-supervised and supervised objectives, any segmentation masks, when available, are binarized and concatenated as auxiliary supervisory signal. 

\subsubsection{Masked Loss}

Each of the loss components \(\mathcal{L}_{sim}\) and \(\mathcal{L}_{smooth}\) are defined over a dense coordinate map. To restrict the loss function to only valid foreground regions (\eg ignore areas of background where coordinate map predictions are meaningless), each loss component may be multiplied point-wise by a binary mask corresponding to foreground regions before aggregation for loss calculation. The procedure used to generate foreground masks in these experiments is detailed below in \ref{Masks}.

\subsection{Implementation Details}

\subsubsection{Network Design} In these experiments, the DeepATLAS network backbone is implemented using a \((64, 64, 64, 1)\) input and a total of five resolution levels (across \(L = 4\) subsamples). The model is initialized with a channel depth of 8 and doubled at each resolution level yielding intermediate backbone feature maps ranging from \((64, 64, 64, 8)\) to \((4, 4, 4, 128)\) in shape. Across all multi-resolution heads, the final model has a total of 4,277,617 trainable model parameters in addition to a 2,097,152 parameter (\(128^3\)) trainable atlas tensor.

\subsubsection{Resampling} \label{Resampling} All downsampling and upsampling operations throughout the model are performed using linear interpolation implemented with an evenly-spaced 3D coordinate grid resampler. To ensure maximum consistency between resampled feature maps at the different resolution levels, the resampler is defined using a reflection invariant, corner-alignment strategy. Given any feature map and its resampled output at a new resolution, a corner-preserving resampling operation ensures that all corners of both feature maps remain the same after resampling. Such a strategy enables the formulation of the multi-resolution heads as residual learners, as a corner-preserving resampling ensures that each resolution is initialized with aligned features from the previous level. 

\subsubsection{Staged Optimization} \label{Optimization} The staged optimization process begins at the deepest \(c_4\) output of shape \((4, 4, 4, 3)\) which is upsampled to the full resolution and used for loss derivation. After convergence, the process is repeated for the next \(c_3\) output of shape \((8, 8, 8, 3)\) and so on, until all resolutions are converged. At each stage only one single head is actively optimized, however due to the residual formulation described above, valid gradients remain for all previous model heads.

Prior to each successive stage of optimization, the \(c_{l}\) head is pretrained to replicate upsampled predictions from the previous resolution \(c_{l+1}\). Given the residual formulation of the multi-resolution heads, this may be accomplished most readily by learning an identity residual function for the current fixed depth feature map \(f_{l} = 0\) such that predictions are completely derived from the upsampled previous resolution fixed depth feature map \(f_{l+1}\). To enforce this behavior, this pretraining step is implemented by freezing all previous model weights except the current resolution head and minimizing the difference between the current \(c_{l}\) and previous upsampled \(c_{l+1}\) coordinate maps.  

\begin{figure*}
    \centering
    \includegraphics[width=\textwidth]{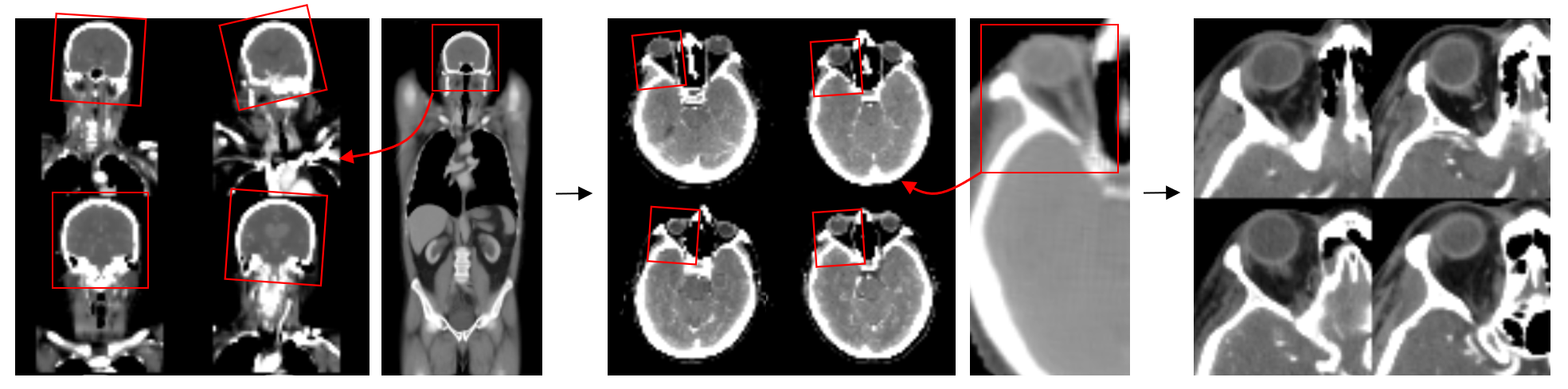}
    \caption{Overview of Cascaded Strategy. To accommodate high-resolution 3D volumes within GPU constraints, a cascaded modeling strategy utilizes fixed input shapes resampled to progressively higher resolutions. In this example, large field-of-view volumes from the head and neck region are resampled to include only more focused views of the head, and subsequently, the orbit. In each step, DeepATLAS model predictions (converged from the prior round of training) are used to project a template bounding cube representing target anatomy defined in the learned atlas space onto all exams (curved red arrow), which in turn are used to produce the next iteration of higher resolution inputs (straight black arrow).}
    \label{fig:cascaded}
\end{figure*}

\subsubsection{Cascaded Model} \label{Cascaded} As DeepATLAS model predictions are inherently dependent on broad contextual information, a 3D model is required for robust performance. However, GPU memory constraints as well as heterogeneity in exam resolution preclude direct optimization on raw full-resolution data. As a result, DeepATLAS implements a cascaded modeling strategy with fixed shape 3D inputs resampled to progressively higher resolutions (Figure \ref{fig:cascaded}). In the first stage of model training, all original data is resampled to a \((64, 64, 64, 1)\) input and used directly for model training. In the subsequent stages of model training, initial predictions are used generate increasingly higher resolution \((64, 64, 64, 1)\) inputs focused on specific target anatomy used for further model training. In this manner, arbitrary increases in prediction resolution may be achieved despite a relative small 3D input shape.

An overview of the iterative cropping procedure is shown in Figure \ref{fig:cascaded}. The key insight is recognizing that the DeepATLAS model itself can be used to project a template bounding cube defined in the learned coordinate space onto any exam (curved red arrow), which in turn can generate resampled data in a standard, uniform field-of-view (straight black arrow). Let \(B_{T} \in \mathbb{R}^3\) represent all points located within a template bounding cube in the learned coordinate space, and \(B_n \in \mathbb{R}^3\) represent all points projected from the template into an exam-specific space using DeepATLAS predictions. While the mapping from \(B_{T} \rightarrow B_n\) is nonlinear, a linear affine transformation matrix \(A\) can be used to approximate this function such that \(B_{T} \cdot A \approx B_n\). Here, \(A\) can be derived explicitly using a least-squares solution to the linear matrix equation, which can in turn be used to project the original vertices of \(B_{T}\). Note that estimating the linear affine matrix in this manner (across all points within \(B_{T}\) instead of just the 8 vertices) is robust to potential outlier predictions.

Though maximum theoretical dataset-specific performance can be achieved by carefully generating crops to each individual target anatomic structure, these experiments balance a tradeoff between performance and efficiency by utilizing up to two serial cropping operations for increased resolution. From initial full field-of-view predictions, iteratively higher resolution predictions are generated for up to six large field-of-views (head, face, neck, chest, abdomen, pelvis) and fourteen small field-of-views (orbit, parotid space, submandibular space, lower neck, mediastinum, cervical spine, thoracic spine, thoracolumbar spine, lumbar spine, right upper quadrant, left upper quadrant, right renal fossa, left renal fossa, lower abdomen), if present in the exam. 

\subsubsection{Pretraining} Prior to staged optimization described above, an optional pretraining procedure may be implemented to improve stability of model convergence. First, a single representative exam is chosen from the training dataset with comprehensive coverage of the target anatomy to seed the initial learnable atlas tensor. In these experiments, the template seed is derived from a single whole-body CT exam. Next, random crops are generated from the single template exam and used to pretrain the most coarse resolution prediction head. Because the location of each random crop is known, this pretraining step may be optimized in a fully supervised manner. 

\subsubsection{Optimization} All experiments are trained using the Adam optimizer \cite{Kingma2014Adam:Optimization} with a batch size of 64 exams. Each individual resolution is optimized for 7,500 iterations yielding a total of 37,500 iterations for all five resolution levels. The learning rate is set to 0.005 at the first resolution and reduced by half after each resolution of training; during optimization, a learning rate decay of 0.995 is applied after every 50 iterations. At the start of each successive resolution of training, learning rate warm-up \cite{Liu2019OnBeyond} is implemented through linear growth of the learning rate from \(10\%\) to \(100\%\) of the target value over 500 iterations. Each experiment is distributed across four NVIDIA Titan RTX 24 GB GDDR6 (Turing architecture) graphics cards.

\subsubsection{Template Segmentation} For one-shot learning experiments, a single reference segmentation of each target anatomic structure is generated within the learned coordinate space. Because the learned template space may vary slightly across each experiment, a standardized method is used to derive reference segmentations without manual tuning. To do so, a model prediction \(\Phi_E = c_n\) is first generated for every example in the training cohort. Then equation \ref{eq:8} is optimized directly by finding the segmentation mask \(M\) (aligned to the atlas \(X\)) that best matches every pair of \(c_n\) predictions and \(m_n\) ground-truth annotations. More formally, the template space segmentation mask \(M\) is simply the volume that minimizes: 

\begin{equation}
    \text{min} \sum_{i=0}^{n} (M \circ c_n - m_n)^2 
\end{equation}

\subsection{Autoencoder} \label{Autoencoder}

\begin{figure}[h!]
    \centering
    \includegraphics[width=\columnwidth]{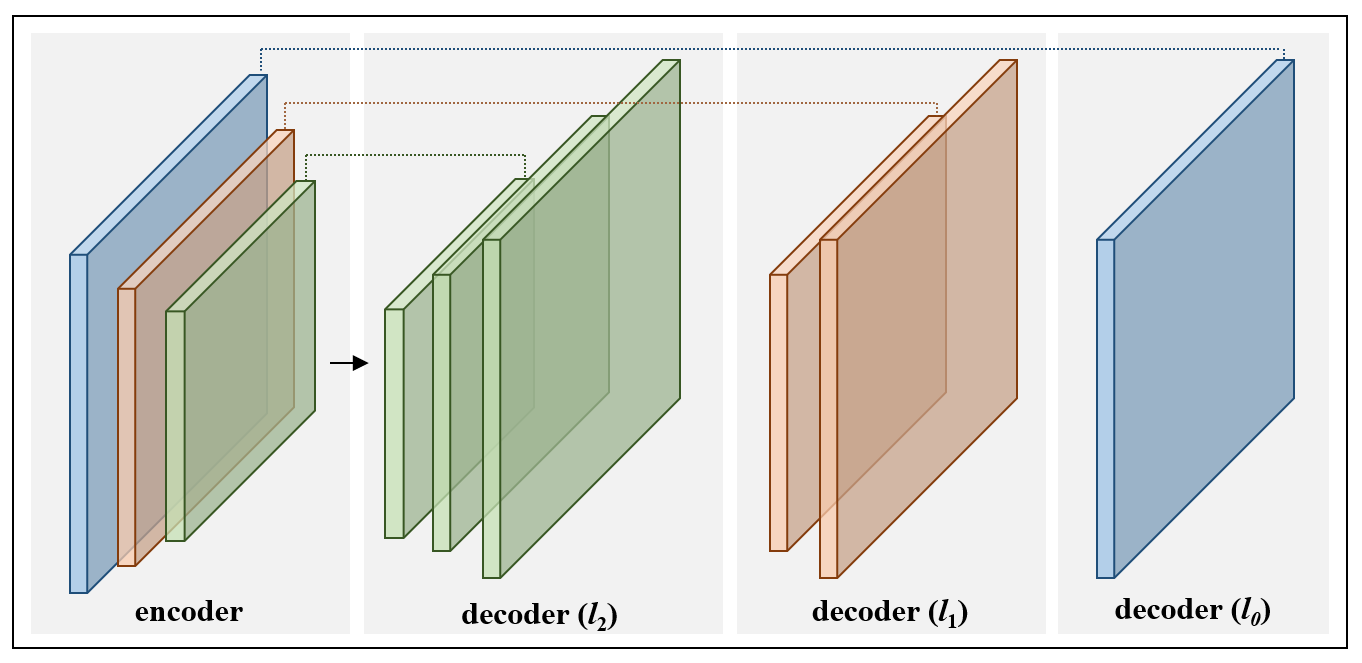}
    \caption{Multi-resolution Autoencoder Heads. Compared to a standard autoencoder which is conditioned to reconstruct only the deepest (most coarse) feature map representation, the proposed model defines separate reconstruction heads for intermediate feature maps at all resolutions, thus encouraging each intermediate learned representation to retain maximum detail.}
    \label{fig:ae}
\end{figure}

The autoencoder for generating feature embeddings used to define the similarity loss is pretrained with multi-resolution reconstruction heads (Figure \ref{fig:ae}). Compared to a standard autoencoder which is conditioned to reconstruct only the deepest (most coarse) feature map representation, the proposed model defines separate reconstruction heads for intermediate feature maps at all resolutions, thus encouraging each intermediate learned representation to retain maximum detail. Without these auxiliary objectives, higher resolution feature maps early in the model need only to retain as much information as may be encoded in the deepest bottleneck layer. Note that maximizing the quality of all intermediate feature maps is critical as multiple feature map resolutions are used in combination to define the similarity loss \(\mathcal{L}_{sim}\) as noted above.

\subsubsection{Feature Maps} By default, the high bit-depth of medical imaging makes it difficult to generate a single, linearly scaled feature embedding with adequate visualization of all tissues types simultaneously due to the wide dynamic range of CT density values (\eg air, fat, soft tissue, bone, contrast material, etc). To address this limitation, the autoencoder reconstruction head may be conditioned to reproduce any arbitrary nonlinear mapping of original CT attenuation values. Doing so encourages the intermediate feature maps to emphasize contrast between desired target tissues. In these experiments, a nonlinear histogram mapping is implemented to maximize overall contrast across a broad range of tissue compositions:

\begin{table}[h!]
    \centering
    \resizebox{\columnwidth}{!}{
        \begin{tblr}{
          column{even} = {c},
          column{3} = {c},
          column{5} = {c},
          column{7} = {c},
          hline{1,3} = {-}{0.08em}}
          
        Raw (Hounsfeld Units) & -256 & -100 & +10  & +50  & +100 & +256 \\
        Normalized            & 0.0  & 0.15 & 0.30 & 0.70 & 0.85 & 1.0  
    
        \end{tblr}
    }
    \label{tab:autoencoder}
\end{table}

Note that this strategy limits the dependence on careful data normalization when training the primary DeepATLAS model, as all relevant target structural detail across a variety of tissue compositions is embedded within the autoencoder-generated feature space. In addition, although the model generates a normalized output from [0, 1], no activation is used in the final autoencoder prediction. This choice is motivated by the observation that a sigmoid (or any other asymptotic) activation heavily skews intermediate representations such that the learned feature space is disproportionately populated by extreme values---if feature map \(F \in (-\infty, +\infty)\) is optimized with a sigmoid activated reconstruction, the vast majority of \(F\) maps to near 0 or 1, and thus any intermediate feature representation is dominated by features at the extreme values. Finally, note that after model training, the corresponding feature maps generated \emph{after} leaky ReLU nonlinearity are extracted for downstream modeling tasks.

\subsubsection{Optimization} In these experiments, the autoencoder is implemented using a \((64, 64, 64, 1)\) resampled input and a total of two multi-resolution heads: original \((64, 64, 64, 1)\) and subsampled \((32, 32, 32, 1)\) resolutions. The architecture is otherwise identical to the main DeepATLAS network backbone including the use of a single residual block at each resolution. During training, all heads are optimized simultaneously. The model is trained using the Adam optimizer \cite{Kingma2014Adam:Optimization} with a batch size of 16 and a total of 20,000 iterations. The learning rate is set to 0.005 with a learning rate decay of 0.995 after every 50 iterations. 

\subsection{Ablation Studies} \label{Ablation Studies}

\subsubsection{Overview}  The following ablation experiments are designed to evaluate the relative contribution between different components of the DeepATLAS loss formulation. For ease of implementation and comparison, only four target anatomic structures across two external datasets are used in these limited experiments, chosen to represent a diversity in tissue composition, body region, and complexity: the mandible (Head-and-Neck Organ-at-Risk; OAR); the right adrenal gland, the right kidney, and the pancreas (Anatomy3; VISCERAL). In addition, only one-shot label propagation is evaluated after pretraining on the full internal cohort (ATLAS-all). 

\subsubsection{Implicit and Explicit Warp Loss}

To assess the relative contribution of the implicit and explicit warp components, the pretraining task is optimized over each loss independently. Table \ref{table:summary-ablation} summarizes the results of one-shot label propagation using either loss component in isolation compared to the baseline strategy. Table \ref{table:ttest-diff-ablation} summarizes the incremental mean differences in performance with each experiment. As expected, optimization using either the explicit loss alone (\(-0.066\) DSC and \(+3.228\) mm HD95, \(p < 0.001\)) or the implicit loss alone (\(-0.043\) DSC and \(+2.473\) mm HD95, \(p < 0.001\)) are significantly worse than the baseline combined approach. Interestingly, the implicit-only model significantly outperforms the explicit-only model (\(+0.022\) DSC and \(-0.755\) mm HD95, \(p < 0.001\)). As discussed in \ref{Loss Formulation}, this observation likely relates to the fact that while both losses promote structural consistency between exams, the implicit warp loss also encourages smooth, invertible coordinate maps.

\subsubsection{Similarity Metric}

To assess the relative benefit of the feature-based similarity metric, the same experiments above are repeated with a standard MSE loss. Tables \ref{table:summary-ablation}-\ref{table:ttest-diff-ablation} summarize the results of this ablation experiment. As expected, optimization using MSE instead of a feature-based metric yields markedly worse performance than the baseline model (\(-0.157\) DSC and \(+5.878\) mm HD95, \(p < 0.001\)). In fact across all other ablation experiments, this single modification yields the worst performance overall (\(p < 0.001\)).

\subsection{Data}

\subsubsection{Internal Cohort}  After local Institutional Review Board approval, all data is downloaded and deidentified per Safe Harbor guidelines including removal of all 18 HIPAA-defined identifiers. For each CT exam, the single axial soft-tissue reconstruction volume with the highest resolution (most number of slices) is included in these experiments. Each volume is converted from original DICOM format to Hierarchial Data Format 5 (HDF5) to facilitate scalable data management. 

\begin{table}[h!]
    \caption{Distribution of Unlabeled Training Data}
    \begin{adjustbox}{max width=\columnwidth,center}
    \begin{tblr}{
      hline{1,13} = {-}{0.08em},
      hline{2,12} = {1-3}{},
    }
    \textbf{Body Part} & \textbf{Count}  & \textbf{Example Protocols}                               &  \\
    head           & 11,436          & CT head with/without contrast, CTA/CTV head, CT perfusion    &  \\
    face           & 1,767           & CT maxillofacial, CT sinus, CT temporal bone                 &  \\
    neck           & 1,036           & CT neck with/without contrast                                &  \\
    head/neck      & 1,156           & CTA head/neck                                                &  \\
    chest          & 11,707          & CT chest with/without contrast, CTA/CTV chest                &  \\
    abdomen/pelvis & 15,466          & CT abdomen/pelvis with/without contrast, CTA abdomen/pelvis  &  \\
    cervical spine & 3,942           & CT cervical spine with/without contrast                      &  \\
    thoracic spine & 1,549           & CT thoracic spine with/without contrast                      &  \\
    lumbar spine   & 1,533           & CT lumbar spine with/without contrast                        &  \\
    whole body     & 1,575           & PET/CT whole body, CTA whole body                            &  \\
    \textit{total} & \textit{51,167} &                                                              &  
    \end{tblr}
    \end{adjustbox}
    \label{tab:data-distribution}
\end{table}

\subsubsection{Preprocessing} For initial training, all original raw data is resampled to a fixed \((64, 64, 64, 1)\) input. This downsampling operation is performed systematically regardless of original data shape or aspect ratio. To maximize signal-to-noise despite aggressive downsampling, a modified box sampling procedure is used: all data is first resampled to an intermediate volume of \((128, 128, 128, 1)\) using linear interpolation followed by an average pooling operation with a \((2, 2, 2)\) kernel. In subsequent stages of training with iteratively higher resolution cropped inputs, a similar strategy is applied to retain high-quality resampled data. 

Prior to training, all data is clipped to a range of [-256, +256] Hounsfeld Units and rescaled to a range of [0, 1]. During optimization, data augmentation is applied dynamically to the input tensors using a 3D affine transformation with scaling (\(70\%\) to \(130\%\)), rotation (-0.5 to +0.5 radians), and translation (-8 to +8 voxels).

\subsubsection{Training Masks} \label{Masks} Binary training masks are applied to the loss function to remove the contribution of non-target regions during optimization. In these experiments, training masks are defined to include all foreground regions (\eg excluding background air). Though these training masks stabilize the optimization process, this strategy is not critical for algorithm convergence. Thus, a simple procedure is adopted to generate approximate training masks in these experiments: (1) apply a Gaussian blur with \(\sigma = 0.7\) to resampled \((64, 64, 64, 1)\) inputs; (2) binarize the data with threshold of -900 Hounsfeld Units; (3) fill all holes; (4) retain largest contiguous 3D region. 

\subsection{Statistics}

\subsubsection{Metrics} All results are reported using approximately half of the labeled external data (N=132) as a fixed test set across all experiments. Spatial overlap between ground-truth \(Y\) and algorithm prediction \(\hat{Y}\) masks is assessed using the Dice score coefficient \cite{Dice1945MeasuresSpecies} (DSC):

\begin{equation}
    \text{DSC} = \frac{2|Y\cap\hat{Y}|}{|Y|+|\hat{Y}|}
\end{equation}

Additionally, the consistency of contour borders is assessed using the symmetric 95th percentile Hausdorff distance \cite{Huttenlocher1993ComparingDistance}. Let \(\text{HD}_{95}(A, B)\) represent the 95th percentile max deviation between all the points in \(A\) to its nearest corresponding point in \(B\). Note that since sets \(A\) and \(B\) may have different sizes, \(\text{HD}_{95}(A, B)\) and \(\text{HD}_{95}(B, A)\) are not necessarily equal in value. Thus, for the set of points along ground-truth \(Y\) and prediction \(\hat{Y}\) masks, the symmetric Hausdorff distance calculation is defined as:

\begin{equation}
    0.5 \cdot (\text{HD}_{95}(Y, \hat{Y}) + \text{HD}_{95}(\hat{Y}, Y))
\end{equation}

Given that a fixed external test set is used throughout all experiments, statistically significant differences in performance between any two experiments is assessed using a paired sample \textit{t}-test across both DSC and HD95 values.

\subsection{Related Work}

\subsubsection{Diffeomorphism} The implicit warp \(\Phi_I\) defines a diffeomorphic function that maps a pair of manifolds from exams \(x_p \rightarrow x_q\) using DeepATLAS coordinate predictions. In a broader context, the estimation of diffeomorphic functions such as \(\Phi\) to align datasets underlies the domain of image registration, and many computational as well as machine learning algorithms have been described for the task \cite{MarkJenkinson2002, Zitova2003ImageSurvey, Oliveira2014MedicalReview, Fu2020DeepReview}. However, an important distinction is that while image registration tasks are formulated to solve \(\Phi\) primarily, DeepATLAS yields only an implicit estimate of \(\Phi\), a secondary calculation defined for pairs of coordinate predictions used only for optimization. Furthermore, while conventional learning-based image registration algorithms optimized with self-supervision may in theory be used for pretraining, the DeepATLAS formulation generates more semantically meaningful representations. Indeed, the registration task is conditioned only to match visually similar patterns without requiring specific knowledge of underlying structure. Additionally, a learning-based registration algorithm produces embeddings that are conditionally assigned only relative to a second reference dataset (which change with each new reference target) and is thus impractical as a pretraining strategy. 

\subsubsection{Automorphism} The explicit warp \(\Phi_E\) defines an automorphic function \(C \rightarrow c_n\) that maps the learned coordinate space and an associated atlas reconstruction to the original input. Standard template spaces, also commonly referenced as atlas spaces, have been previously derived for various anatomy most notably in neuroimaging such as the Talairach \cite{Talairach1988Co-planarImaging} and Montreal Neurological Institute (MNI; \cite{Collins1994AutomaticSpace.}) coordinate systems. Conventionally, these coordinate spaces are projected onto individual exams through manual registration of a corresponding template atlas. By contrast, a forward-pass of the DeepATLAS model directly labels each position with its standard coordinate using the raw exam alone. Indeed, while a reconstructed atlas corresponding to the learned coordinate space is generated as an artifact of pretraining, neither the atlas nor any other external dependency is needed for model prediction.

Thus, while atlas-based registration methods require similarity to a predefined external template, DeepATLAS leverages a more flexible internal learned representation. This seemingly simple modification nonetheless has several important implications. First, DeepATLAS is broadly relevant to domains beyond which a well-defined atlas currently exists. More critically, DeepATLAS generalizes to problems that are difficult to solve using conventional image registration alone. For example, image registration of a whole body atlas to any single (partial field-of-view) exam is poorly defined because at baseline it is unknown which portions of the atlas overlap with the exam (and which portions should be ignored). Along similar lines, conventional atlas registration fails when any single template is insufficient to capture underlying data heterogeneity, as may occur when a cohort comprises of multiple phases of contrast administration or other variations in imaging protocol.

\clearpage

\clearpage

\begin{table*}[t]
    \caption{Summary of Dice Score Performance}
    \label{table:summary-dsc}
    \begin{adjustbox}{width=\textwidth}
    \begin{tblr}{
      column{2} = {c},
      column{3} = {c},
      column{4} = {c},
      column{5} = {c},
      column{7} = {c},
      column{9} = {c},
      column{10} = {c},
      cell{1}{2} = {c=4}{},
      cell{1}{9} = {c=2}{},
      hline{1,62} = {-}{0.08em},
      hline{2} = {2-5,7,9-10}{},
      hline{3-4,25-26,32-33,40-41} = {-}{},
    }
    ~                                   & Unsupervised (One-Shot Propagation)   &        &         &         & ~ & Semisupervised & ~ & Supervised   &              \\
                                        & ATLAS-500                             & ATLAS-5k & ATLAS-all & ATLAS-ext &   & joint loss    &   & pretrain (+) & pretrain (-) \\
    Head-and-Neck Organ-at-Risk (OAR)   &                                       &        &         &         &   &                &   &              &              \\
    ~ brain                             & 0.959                                 & 0.974  & 0.979   & 0.976   & ~ & 0.975          & ~ & 0.980        & 0.964        \\
    ~ brainstem                         & 0.796                                 & 0.839  & 0.855   & 0.849   &   & 0.851          &   & 0.806        & 0.788        \\
    ~ cochlea, left                     & 0.264                                 & 0.410  & 0.466   & 0.550   &   & 0.608          &   & 0.000        & 0.000        \\
    ~ cochlea, right                    & 0.188                                 & 0.341  & 0.425   & 0.485   &   & 0.483          &   & 0.000        & 0.000        \\
    ~ lacrimal gland, left              & 0.361                                 & 0.542  & 0.611   & 0.639   &   & 0.652          &   & 0.548        & 0.500        \\
    ~ lacrimal gland, right             & 0.458                                 & 0.549  & 0.606   & 0.640   &   & 0.657          &   & 0.598        & 0.525        \\
    ~ lens, left                        & 0.281                                 & 0.487  & 0.556   & 0.694   &   & 0.707          &   & 0.481        & 0.104        \\
    ~ lens, right                       & 0.304                                 & 0.461  & 0.604   & 0.714   &   & 0.716          &   & 0.511        & 0.070        \\
    ~ lung, left                        & 0.811                                 & 0.873  & 0.914   & 0.932   &   & 0.928          &   & 0.935        & 0.829        \\
    ~ lung, right                       & 0.832                                 & 0.900  & 0.928   & 0.946   &   & 0.937          &   & 0.947        & 0.906        \\
    ~ mandible                          & 0.464                                 & 0.637  & 0.818   & 0.848   &   & 0.858          &   & 0.866        & 0.839        \\
    ~ optic nerve, left                 & 0.439                                 & 0.549  & 0.624   & 0.694   &   & 0.693          &   & 0.641        & 0.388        \\
    ~ optic nerve, right                & 0.445                                 & 0.524  & 0.613   & 0.687   &   & 0.700          &   & 0.619        & 0.405        \\
    ~ orbit, left                       & 0.751                                 & 0.834  & 0.877   & 0.909   &   & 0.909          &   & 0.882        & 0.848        \\
    ~ orbit, right                      & 0.773                                 & 0.850  & 0.879   & 0.907   &   & 0.908          &   & 0.881        & 0.855        \\
    ~ parotid gland, left               & 0.649                                 & 0.706  & 0.732   & 0.765   &   & 0.772          &   & 0.778        & 0.745        \\
    ~ parotid gland, right              & 0.719                                 & 0.753  & 0.776   & 0.787   &   & 0.770          &   & 0.773        & 0.735        \\
    ~ spinal canal                      & 0.734                                 & 0.824  & 0.864   & 0.896   &   & 0.895          &   & 0.865        & 0.820        \\
    ~ submandibular gland, left         & 0.595                                 & 0.660  & 0.661   & 0.704   &   & 0.735          &   & 0.722        & 0.651        \\
    ~ submandibular gland, right        & 0.605                                 & 0.666  & 0.686   & 0.689   & ~ & 0.695          & ~ & 0.705        & 0.646        \\
    \textit{~ average}                  & 0.571                                 & 0.669  & 0.724   & 0.766   & ~ & 0.773          & ~ & 0.677        & 0.581        \\
    CT Multi-Organ (CT-ORG)             &                                       &        &         &         &   &                &   &              &              \\
    ~ bladder                           & 0.556                                 & 0.569  & 0.570   & 0.590   &   & 0.619          &   & 0.792        & 0.646        \\
    ~ kidney, left                      & 0.714                                 & 0.813  & 0.835   & 0.846   &   & 0.845          &   & 0.871        & 0.835        \\
    ~ kidney, right                     & 0.720                                 & 0.806  & 0.819   & 0.827   &   & 0.829          &   & 0.858        & 0.798        \\
    ~ liver                             & 0.834                                 & 0.880  & 0.885   & 0.899   &   & 0.906          &   & 0.931        & 0.899        \\
    ~ lung                              & 0.846                                 & 0.917  & 0.932   & 0.927   & ~ & 0.926          & ~ & 0.945        & 0.929        \\
    \textit{~ average}                  & 0.734                                 & 0.797  & 0.808   & 0.818   &   & 0.825          &   & 0.880        & 0.821        \\
    Chest Organ-at-Risk (StructSeg)     &                                       &        &         &         &   &                &   &              &              \\
    ~ esophagus                         & 0.522                                 & 0.593  & 0.620   & 0.646   &   & 0.662          &   & 0.555        & 0.410        \\
    ~ heart                             & 0.882                                 & 0.921  & 0.926   & 0.927   &   & 0.927          &   & 0.910        & 0.896        \\
    ~ lung, left                        & 0.850                                 & 0.917  & 0.930   & 0.937   &   & 0.935          &   & 0.945        & 0.941        \\
    ~ lung, right                       & 0.885                                 & 0.937  & 0.948   & 0.954   &   & 0.953          &   & 0.954        & 0.949        \\
    ~ spinal canal                      & 0.726                                 & 0.825  & 0.860   & 0.885   &   & 0.895          &   & 0.844        & 0.798        \\
    ~ trachea                           & 0.609                                 & 0.716  & 0.737   & 0.786   & ~ & 0.783          & ~ & 0.769        & 0.697        \\
    \textit{~ average}                  & 0.746                                 & 0.818  & 0.837   & 0.856   &   & 0.859          &   & 0.830        & 0.782        \\
    Anatomy3 (VISCERAL)                 &                                       &        &         &         &   &                &   &              &              \\
    ~ adrenal, left                     & 0.222                                 & 0.371  & 0.350   & 0.433   &   & 0.411          &   & 0.120        & 0.121        \\
    ~ adrenal, right                    & 0.149                                 & 0.319  & 0.349   & 0.390   &   & 0.387          &   & 0.000        & 0.000        \\
    ~ aorta                             & 0.701                                 & 0.777  & 0.801   & 0.826   &   & 0.837          &   & 0.807        & 0.724        \\
    ~ bladder                           & 0.595                                 & 0.647  & 0.657   & 0.694   &   & 0.739          &   & 0.832        & 0.713        \\
    ~ gallbladder                       & 0.197                                 & 0.279  & 0.286   & 0.341   &   & 0.355          &   & 0.200        & 0.166        \\
    ~ kidney, left                      & 0.761                                 & 0.869  & 0.887   & 0.910   &   & 0.908          &   & 0.913        & 0.852        \\
    ~ kidney, right                     & 0.766                                 & 0.871  & 0.883   & 0.905   &   & 0.906          &   & 0.923        & 0.836        \\
    ~ l1 vertebrae                      & 0.557                                 & 0.622  & 0.617   & 0.802   &   & 0.803          &   & 0.618        & 0.515        \\
    ~ liver                             & 0.858                                 & 0.912  & 0.919   & 0.933   &   & 0.934          &   & 0.940        & 0.915        \\
    ~ lung, left                        & 0.892                                 & 0.946  & 0.955   & 0.958   &   & 0.957          &   & 0.963        & 0.954        \\
    ~ lung, right                       & 0.907                                 & 0.955  & 0.960   & 0.963   &   & 0.962          &   & 0.967        & 0.956        \\
    ~ pancreas                          & 0.467                                 & 0.539  & 0.541   & 0.623   &   & 0.632          &   & 0.545        & 0.456        \\
    ~ psoas muscle, left                & 0.718                                 & 0.792  & 0.801   & 0.823   &   & 0.827          &   & 0.802        & 0.701        \\
    ~ psoas muscle, right               & 0.688                                 & 0.771  & 0.773   & 0.800   &   & 0.821          &   & 0.791        & 0.698        \\
    ~ rectus abdominal, left            & 0.359                                 & 0.572  & 0.637   & 0.627   &   & 0.595          &   & 0.640        & 0.553        \\
    ~ rectus abdominal, right           & 0.339                                 & 0.573  & 0.635   & 0.645   &   & 0.590          &   & 0.588        & 0.467        \\
    ~ spleen                            & 0.807                                 & 0.878  & 0.878   & 0.928   &   & 0.929          &   & 0.920        & 0.840        \\
    ~ sternum                           & 0.580                                 & 0.709  & 0.741   & 0.770   &   & 0.772          &   & 0.328        & 0.085        \\
    ~ thyroid                           & 0.456                                 & 0.490  & 0.541   & 0.517   &   & 0.564          &   & 0.482        & 0.403        \\
    ~ trachea                           & 0.684                                 & 0.774  & 0.825   & 0.883   & ~ & 0.884          & ~ & 0.870        & 0.804        \\
    ~\textit{ average}                  & 0.585                                 & 0.683  & 0.702   & 0.739   & ~ & 0.741          & ~ & 0.662        & 0.588        
    \end{tblr}
    \end{adjustbox}
\end{table*}

\clearpage

\begin{table*}[t]
    \caption{Summary of 95th Percentile Hausdorff Distance Performance}
    \label{table:summary-hd}
    \begin{adjustbox}{width=\textwidth}
    \begin{tblr}{
      column{2} = {c},
      column{3} = {c},
      column{4} = {c},
      column{5} = {c},
      column{7} = {c},
      column{9} = {c},
      column{10} = {c},
      cell{1}{2} = {c=4}{},
      cell{1}{9} = {c=2}{},
      hline{1,62} = {-}{0.08em},
      hline{2} = {2-5,7,9-10}{},
      hline{3-4,25-26,32-33,40-41} = {-}{},
    }
    ~                                 & Unsupervised (One-Shot Propagation) &        &         &         & ~ & Semisupervised & ~ & Supervised   &              \\
                                      & ATLAS-500                           & ATLAS-5k & ATLAS-all & ATLAS-ext &   & joint loss    &   & pretrain (+) & pretrain (-) \\
    Head-and-Neck Organ-at-Risk (OAR) &                                     &        &         &         &   &                &   &              &              \\
    ~ brain                           & 4.352                               & 3.208  & 2.695   & 3.013   &   & 3.075          &   & 2.814        & 5.147        \\
    ~ brainstem                       & 5.174                               & 4.017  & 3.641   & 3.777   &   & 3.795          &   & 5.834        & 6.418        \\
    ~ cochlea, left                   & 3.570                               & 2.462  & 1.803   & 1.673   &   & 1.594          &   & n/a          & n/a          \\
    ~ cochlea, right                  & 3.892                               & 2.790  & 2.564   & 2.303   &   & 2.371          &   & n/a          & n/a          \\
    ~ lacrimal gland, left            & 4.916                               & 3.865  & 3.496   & 3.227   &   & 3.079          &   & 3.757        & 3.904        \\
    ~ lacrimal gland, right           & 4.363                               & 3.604  & 3.263   & 3.028   &   & 2.929          &   & 3.235        & 3.669        \\
    ~ lens, left                      & 4.198                               & 2.940  & 2.591   & 2.000   &   & 1.738          &   & 2.431        & 3.552        \\
    ~ lens, right                     & 4.179                               & 3.199  & 2.428   & 1.831   &   & 1.776          &   & 2.374        & 3.764        \\
    ~ lung, left                      & 19.251                              & 14.646 & 9.476   & 8.963   &   & 9.062          &   & 13.201       & 22.921       \\
    ~ lung, right                     & 19.154                              & 11.446 & 8.683   & 6.428   &   & 8.036          &   & 13.120       & 16.531       \\
    ~ mandible                        & 11.429                              & 6.945  & 3.961   & 3.707   &   & 3.421          &   & 3.046        & 4.075        \\
    ~ optic nerve, left               & 4.359                               & 5.600  & 3.306   & 2.944   &   & 2.933          &   & 4.747        & 7.826        \\
    ~ optic nerve, right              & 4.178                               & 5.170  & 2.824   & 2.613   &   & 2.637          &   & 4.107        & 7.297        \\
    ~ orbit, left                     & 4.195                               & 2.830  & 2.325   & 2.003   &   & 2.441          &   & 2.703        & 2.982        \\
    ~ orbit, right                    & 3.979                               & 2.682  & 2.375   & 2.086   &   & 2.512          &   & 2.652        & 2.885        \\
    ~ parotid gland, left             & 12.067                              & 10.952 & 11.140  & 9.937   &   & 8.655          &   & 8.410        & 9.552        \\
    ~ parotid gland, right            & 8.593                               & 7.710  & 7.683   & 7.670   &   & 7.671          &   & 7.251        & 8.764        \\
    ~ spinal canal                    & 15.870                              & 13.901 & 13.526  & 13.256  &   & 13.463         &   & 13.723       & 22.264       \\
    ~ submandibular gland, left       & 7.167                               & 6.002  & 5.840   & 5.571   &   & 5.125          &   & 5.125        & 6.268        \\
    ~ submandibular gland, right      & 6.528                               & 5.798  & 5.294   & 5.387   & ~ & 5.483          & ~ & 5.189        & 5.946        \\
    \textit{~ average}                & 7.571                               & 5.988  & 4.946   & 4.571   & ~ & 4.590          & ~ & 5.762        & 7.987        \\
    CT Multi-Organ (CT-ORG)           &                                     &        &         &         &   &                &   &              &              \\
    ~ bladder                         & 23.929                              & 24.683 & 24.527  & 23.179  &   & 21.684         &   & 11.357       & 19.058       \\
    ~ kidney, left                    & 17.513                              & 14.034 & 13.816  & 15.589  &   & 13.940         &   & 14.302       & 14.526       \\
    ~ kidney, right                   & 19.200                              & 15.373 & 15.754  & 15.454  &   & 15.501         &   & 15.237       & 16.244       \\
    ~ liver                           & 24.127                              & 21.082 & 21.511  & 20.381  &   & 18.485         &   & 14.214       & 18.998       \\
    ~ lung                            & 33.179                              & 26.751 & 25.348  & 26.101  & ~ & 25.626         & ~ & 23.621       & 24.685       \\
    \textit{~ average}                & 23.590                              & 20.385 & 20.191  & 20.141  &   & 19.047         &   & 15.746       & 18.702       \\
    Chest Organ-at-Risk (StructSeg)   &                                     &        &         &         &   &                &   &              &              \\
    ~ esophagus                       & 12.851                              & 11.402 & 11.096  & 10.595  &   & 9.650          &   & 15.145       & 29.625       \\
    ~ heart                           & 10.697                              & 8.358  & 7.708   & 7.972   &   & 8.361          &   & 9.538        & 9.947        \\
    ~ lung, left                      & 18.450                              & 9.107  & 8.197   & 7.882   &   & 7.760          &   & 5.810        & 6.602        \\
    ~ lung, right                     & 16.560                              & 8.819  & 7.389   & 6.398   &   & 6.836          &   & 5.480        & 5.911        \\
    ~ spinal canal                    & 5.465                               & 3.918  & 2.978   & 2.906   &   & 2.285          &   & 4.951        & 5.905        \\
    ~ trachea                         & 13.457                              & 10.782 & 10.789  & 10.388  & ~ & 10.360         & ~ & 12.131       & 13.078       \\
    \textit{~ average}                & 12.913                              & 8.731  & 8.026   & 7.690   &   & 7.542          &   & 8.843        & 11.845       \\
    Anatomy3 (VISCERAL)               &                                     &        &         &         &   &                &   &              &              \\
    ~ adrenal, left                   & 14.834                              & 12.704 & 12.377  & 10.576  &   & 10.338         &   & 12.130       & 11.106       \\
    ~ adrenal, right                  & 16.416                              & 11.767 & 11.295  & 9.493   &   & 9.044          &   & n/a          & n/a          \\
    ~ aorta                           & 15.249                              & 12.575 & 11.747  & 10.799  &   & 10.050         &   & 11.448       & 13.674       \\
    ~ bladder                         & 22.811                              & 18.192 & 18.125  & 16.799  &   & 15.362         &   & 10.329       & 15.840       \\
    ~ gallbladder                     & 23.855                              & 19.854 & 20.126  & 18.185  &   & 18.017         &   & 19.534       & 24.641       \\
    ~ kidney, left                    & 16.185                              & 9.968  & 9.439   & 6.999   &   & 7.403          &   & 9.410        & 11.411       \\
    ~ kidney, right                   & 14.618                              & 9.476  & 8.725   & 6.925   &   & 6.700          &   & 6.755        & 11.842       \\
    ~ l1 vertebrae                    & 13.384                              & 12.200 & 12.850  & 6.078   &   & 6.257          &   & 11.804       & 14.241       \\
    ~ liver                           & 16.999                              & 11.281 & 10.538  & 8.922   &   & 8.411          &   & 7.644        & 11.403       \\
    ~ lung, left                      & 14.323                              & 6.293  & 4.928   & 4.759   &   & 4.829          &   & 4.055        & 4.982        \\
    ~ lung, right                     & 13.075                              & 5.650  & 4.923   & 4.720   &   & 4.718          &   & 3.824        & 5.183        \\
    ~ pancreas                        & 20.726                              & 19.847 & 19.742  & 17.443  &   & 17.425         &   & 18.822       & 20.688       \\
    ~ psoas muscle, left              & 17.034                              & 13.563 & 13.321  & 11.527  &   & 11.558         &   & 15.592       & 20.360       \\
    ~ psoas muscle, right             & 19.066                              & 15.618 & 15.429  & 14.609  &   & 12.593         &   & 17.727       & 23.399       \\
    ~ rectus abdominal, left          & 28.151                              & 22.658 & 19.820  & 23.579  &   & 28.370         &   & 25.019       & 23.895       \\
    ~ rectus abdominal, right         & 29.192                              & 21.161 & 17.951  & 21.072  &   & 23.927         &   & 26.814       & 30.460       \\
    ~ spleen                          & 13.638                              & 8.903  & 8.911   & 5.118   &   & 6.090          &   & 5.511        & 12.520       \\
    ~ sternum                         & 18.787                              & 15.051 & 13.537  & 12.595  &   & 12.953         &   & 29.175       & 44.628       \\
    ~ thyroid                         & 13.257                              & 13.880 & 11.298  & 10.052  &   & 9.220          &   & 10.178       & 12.352       \\
    ~ trachea                         & 8.076                               & 6.697  & 5.345   & 3.301   & ~ & 3.017          & ~ & 3.304        & 5.137        \\
    ~\textit{ average}                & 17.484                              & 13.367 & 12.521  & 11.178  & ~ & 11.314         & ~ & 13.109       & 16.724       
    \end{tblr}
    \end{adjustbox}
\end{table*}

\clearpage

\begin{table*}[t]
    \caption{Ablation Experiments - Summary of Dice Score and 95th Percentile Hausdorff Distance Performance}
    \label{table:summary-ablation}
    \centering
    \begin{tblr}{
      row{3} = {c},
      row{9} = {c},
      column{even} = {c},
      column{3} = {c},
      column{5} = {c},
      column{7} = {c},
      cell{1}{4} = {c=2}{},
      cell{3}{1} = {c=7}{},
      cell{9}{1} = {c=7}{},
      hline{1,15} = {-}{0.08em},
      hline{2} = {2,4-5,7}{},
      hline{3-4,9-10} = {-}{},
    }
    ~                         & Baseline & ~ & Warp Lass     &               & ~ & Similarity Loss \\
                              & One-Shot &   & Explicit-Only & Implicit-Only &   & MAE             \\
    Dice Score                &          &   &               &               &   &                 \\
    mandible                  & 0.848    &   & 0.804         & 0.846         &   & 0.800           \\
    adrenal, right            & 0.390    &   & 0.315         & 0.345         &   & 0.181           \\
    kidney, right             & 0.905    &   & 0.865         & 0.869         &   & 0.700           \\
    pancreas                  & 0.623    & ~ & 0.520         & 0.536         & ~ & 0.465           \\
    \textit{average}          & 0.691    &   & 0.626         & 0.649         &   & 0.537           \\
    Hausdorff Distance (mm)   &          &   &               &               &   &                 \\
    mandible                  & 3.707    &   & 4.119         & 3.802         &   & 4.365           \\
    adrenal, right            & 9.493    &   & 13.393        & 11.679        &   & 16.617          \\
    kidney, right             & 6.925    &   & 11.600        & 11.765        &   & 18.534          \\
    pancreas                  & 17.443   & ~ & 21.066        & 19.810        & ~ & 20.845          \\
    \textit{average}          & 9.392    & ~ & 12.545        & 11.764        & ~ & 15.090          
    \end{tblr}
\end{table*}

\begin{table*}[t]
    \caption{Ablation Experiments - Pairwise Differences in Dice Score and Hausdorff Distance}
    \label{table:ttest-diff-ablation}
    \centering
    \begin{tblr}{
      row{3} = {c},
      row{8} = {c},
      column{even} = {c},
      column{3} = {c},
      column{5} = {c},
      column{7} = {c},
      cell{1}{4} = {c=2}{},
      cell{3}{1} = {c=7}{},
      cell{8}{1} = {c=7}{},
      hline{1,13} = {-}{0.08em},
      hline{2} = {2,4-5,7}{},
      hline{3-4,8-9} = {-}{},
    }
    ~                         & Baseline                   & ~ & Warp Lass                  &                            & ~ & Similarity Loss            \\
    ~                         & One-Shot                   & ~ & Explicit-Only              & Implicit-Only              & ~ & MAE                        \\
    Dice Score                &                            &   &                            &                            &   &                            \\
    One-Shot                  & -                          &   & \(-0.066\) \((p < 0.001)\) & \(-0.043\) \((p < 0.001)\) &   & \(-0.157\) \((p < 0.001)\) \\
    Explicit-Only             & \(+0.066\) \((p < 0.001)\) &   & -                          & \(+0.022\) \((p < 0.001)\) &   & \(-0.091\) \((p < 0.001)\) \\
    Implicit-Only             & \(+0.043\) \((p < 0.001)\) &   & \(-0.022\) \((p < 0.001)\) & -                          &   & \(-0.114\) \((p < 0.001)\) \\
    MAE                       & \(+0.157\) \((p < 0.001)\) & ~ & \(+0.091\) \((p < 0.001)\) & \(+0.114\) \((p < 0.001)\) & ~ & -                          \\
    Hausdorff Distance (mm)   &                            &   &                            &                            &   &                            \\
    One-Shot                  & -                          &   & \(+3.228\) \((p < 0.001)\) & \(+2.473\) \((p < 0.001)\) &   & \(+5.878\) \((p < 0.001)\) \\
    Explicit-Only             & \(-3.228\) \((p < 0.001)\) &   & -                          & \(-0.755\) \((p = 0.037)\) &   & \(+2.650\) \((p < 0.001)\) \\
    Implicit-Only             & \(-2.473\) \((p < 0.001)\) &   & \(+0.755\) \((p = 0.037)\) & -                          &   & \(+3.405\) \((p < 0.001)\) \\
    MAE                       & \(-5.878\) \((p < 0.001)\) & ~ & \(-2.650\) \((p < 0.001)\) & \(-3.405\) \((p < 0.001)\) & ~ & -
    \end{tblr}
\end{table*}

\end{document}